\documentclass[10pt,twocolumn,letterpaper]{article}

\usepackage[pagenumbers]{cvpr} 

\usepackage{times}
\usepackage{epsfig}
\usepackage{graphicx}
\usepackage{amsmath}
\usepackage{amssymb}

\usepackage{multirow}
\usepackage{booktabs}
\usepackage{subcaption}
 \usepackage{hhline}
 \usepackage{colortbl}
 \usepackage{graphics} 
\usepackage{mathptmx} 
\usepackage{makecell}
\usepackage[T1]{fontenc}
\usepackage{bbding}
\usepackage{wasysym}

\usepackage{colortbl}
\usepackage{mathrsfs}
\usepackage{booktabs}
\usepackage{footmisc}
\usepackage{tablefootnote}

\usepackage{amsmath}
\usepackage{algpseudocode}
\usepackage{algorithm}
\algnewcommand{\IIf}[1]{\State\algorithmicif\ #1\ \algorithmicthen}
\algnewcommand{\EndIIf}{\unskip\ \algorithmicend\ \algorithmicif}
\usepackage{algcompatible}
\usepackage{amssymb}
\usepackage{pifont}
\newcommand{\xmark}{\ding{55}}%
\newcommand{\mypar}[1]{\vspace{0.1cm}\noindent\textbf{#1}.}
\usepackage{balance}

\usepackage{footnote}

\usepackage[pagebackref=true,breaklinks=true,letterpaper=true,colorlinks,bookmarks=false]{hyperref}
\hypersetup{colorlinks=true,linkcolor=red,citecolor=blue,urlcolor=black}

\usepackage[capitalize]{cleveref}
\crefname{section}{Sec.}{Secs.}
\Crefname{section}{Section}{Sections}
\Crefname{table}{Table}{Tables}
\crefname{table}{Tab.}{Tabs.}

\begin{document}

\title{\emph{Should I take a walk?} 
Estimating Energy Expenditure from Video Data}

\author{Kunyu Peng$^{*}$, Alina Roitberg$^{*}$, Kailun Yang, Jiaming Zhang, and Rainer Stiefelhagen\\
$^{*}$denotes equal contribution\\
Karlsruhe Institute of Technology\\
{\tt\small firstname.lastname@kit.edu}
}
\maketitle

\begin{abstract}

We explore the problem of automatically inferring the amount of kilocalories used by human during physical activity from his/her video observation. To study this underresearched task, we introduce \textsl{Vid2Burn} -- an omni-source benchmark for estimating caloric expenditure from video data featuring both, high- and low-intensity activities for which we derive energy expenditure annotations based on models established in medical literature.
 
In practice, a training set would only cover a certain amount of activity types, and it is important to validate, if the model indeed captures the essence of energy expenditure, (\eg, how many and which muscles are involved and how intense they work) instead of memorizing fixed values of specific activity categories seen during training. Ideally, the models should look beyond such category-specific biases and regress the caloric cost in videos depicting activity categories not explicitly present during training. With this property in mind, \textsl{Vid2Burn} is accompanied with a cross-category benchmark, where the task is to regress caloric expenditure for types of physical activities not present during training. An extensive evaluation of state-of-the-art approaches for video recognition modified for the energy expenditure estimation task demonstrates the difficulty of this problem, especially for new activity types at test-time, marking a new research direction. Dataset and code are available at \url{https://github.com/KPeng9510/Vid2Burn}\footnote[1]{Acknowledgments: The research leading to these results was supported by the SmartAge
project sponsored by the Carl Zeiss Stiftung (P2019-01-003; 2021-2026).}. 
\end{abstract}

\section{Introduction}
\begin{figure}
    \centering
    \includegraphics[width= \linewidth]{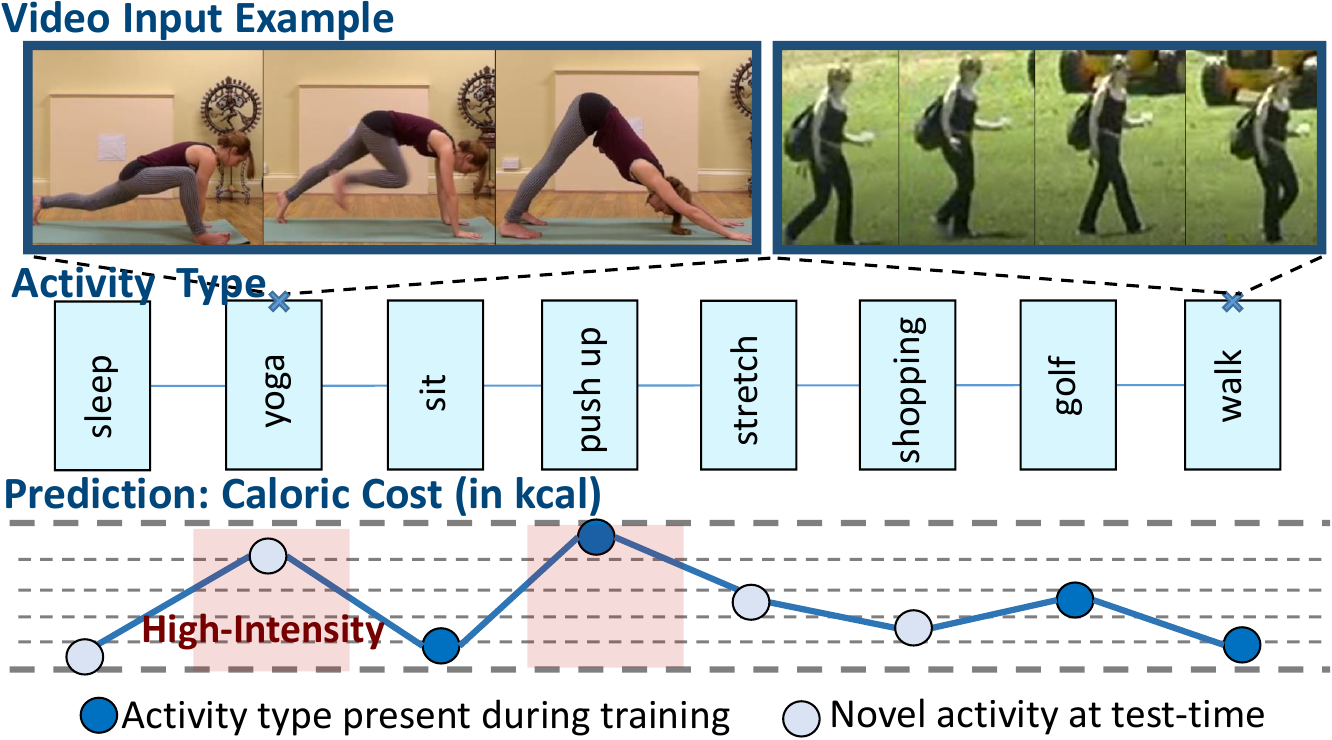}
    \vskip-1ex
    \caption{
   Our goal is to determine the amount of kilocalories burned during physical activity from video observation.
   To this intent, we collect \textsl{Vid2Burn} -- an omni-source dataset with videos of people engaged in low- and high intensity activities and caloric cost annotations derived from established physiological models.
   }
   \vskip-4ex
\label{fig:intro}
\end{figure}

If you would ask people to honestly answer ``\textit{Why do you go to the gym?}'' a frequent reply would be to burn calories.
Physical activity is connected with our health and is an important element in prevention of obesity, diabetes or high blood pressure\footnote[2]{World Health Organization (WHO) - Physical inactivity a leading cause of disease and disability: \url{https://www.who.int/news/item/04-04-2002-physical-inactivity-a-leading-cause-of-disease-and-disability-warns-who}} -- issues which are amplified through the recent Covid-19 lockdowns and the home office regulations~\cite{amini2021covid}.
With the rise of health tracking apps, automatic inference of energy expenditure is rapidly gaining attention~\cite{albinali2010using,barut2020multitask,nakamura2017jointly,masullo2018calorinet,wang2018calorific, zhang2021energy}, but almost all prior research has focused on signals obtained from wearable devices, such as smart watches or heart rate monitoring chest straps.
While such sensors are not always present at hand or comfortable to wear, most people can easily access a video camera in their phone or laptop.
Apart from helping the users interested in tracking their exercise and maintaining active lifestyle, recent studies in gerontology highlight the benefits of automatically tracking the level of physical activity in assistive smart homes in order to support the elderly~\cite{reeder2020older, jo2021elderly,awais2018physical}.

As important as it is for our health, understanding physical activity offers new technical challenges in computer vision.
Excellent progress has been made in the field of human activity recognition~\cite{qiu2017learning,tran2015learning,xie2018rethinking,feichtenhofer2017spatiotemporal,carreira2017quo} with remarkable accuracies reported on datasets such as HMDB-51~\cite{kuehne2011hmdb} or Kinetics~\cite{carreira2017quo}.
However, when facing our task of estimating caloric expenditure from human observations, these methods will face two main obstacles.
First, the cornerstone of past research lies in rather rigid \textit{categorization} into predefined actions. 
These categories are often relatively coarse, (\eg, ``football'' vs. ``jogging''),
so that the scene context provides the network with an excellent shortcut to the decision, leaving the actual moving person behind~\cite{choi2019can,weinzaepfel2021mimetics}.
Our task however requires \textit{fine-grained} understanding of human movement, as medical research~\cite{caspersen1985physical} lists \textit{which muscles are active} and \textit{how hard they work} as the main drivers of energy expenditure (although a multitude of further factors influence this complex physiological process).
\begin{figure*}[t]
\begin{center}
\includegraphics[width = 1\textwidth]{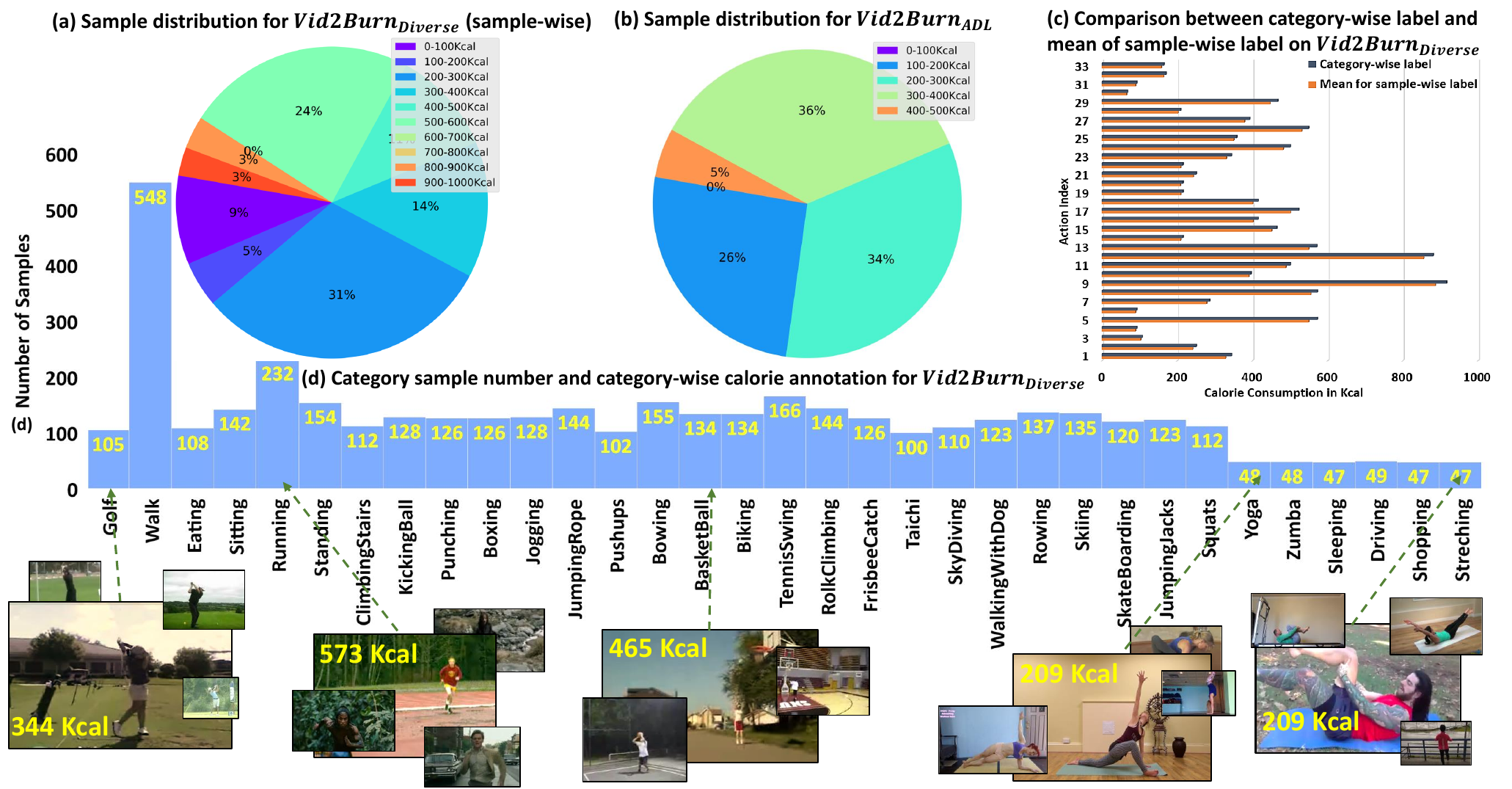}
\end{center}
\vskip-3ex
\caption{An overview of the dataset statistics. The statistics of the caloric cost values are summarized as pie charts in (a) and (b) for \textsl{Vid2Burn\textsubscript{Diverse}} and \textsl{Vid2Burn\textsubscript{ADL}} respectively. The caloric cost annotation statistics grouped by the individual activity categories are provided in (c), where blue bars represent the category-wise annotations, orange bars stand for the average of the sample-wise values and the bar index on the vertical axis indicates the action ID (which correspond to the order of activities listed below the blue histogram). 
The sample frequencies for different action categories in \textsl{Vid2Burn\textsubscript{Diverse}} are visualized in the blue bar chart (d) with multiple visual examples. }
\label{fig:diverse}
\vskip-3ex
\end{figure*} 
A second key challenge is  to encourage the model to  capture the essence of energy expenditure instead of memorizing average values of specific activity categories seen during training.
Deep neural networks are prone to learning shortcuts~\cite{hendricks2018women, geirhos2020shortcut, choi2019can} and  internally casting the calorie regression problem as an ``easier'' task of activity categorization which might be one of such potential shortcuts.
Even if the annotations are continuous calorie values and not rigid categories, in practice, the training set can only cover a finite amount of activity types. 
Ideally, our model should not be bounded to category-specific biases and indeed learn the nature of activity-induced energy expenditure by, \eg, understanding the type and intensity of bodily movement produced by the skeletal muscles.
When developing an energy expenditure benchmark, it is therefore critical to evaluate the results on types of physical activity not present during training.

In this paper, we explore the new research direction of inferring activity-induced caloric cost by observing the human in video, as shown in Fig.~\ref{fig:intro}.
To tackle the lack of public large-scale datasets, we introduce \textsl{Vid2Burn} - a new omni-source benchmark spanning $9789$ video examples of people engaged in different activities with corresponding annotations designed based on models established in medical literature~\cite{keytel2005prediction} from (1) current activity category (2) intensity of the skeleton movement and (3) heart rate measurements obtained for a subset of activities (household activities) in a complementary study.
Videos in the dataset are chosen from four diverse activity recognition datasets~\cite{soomro2012ucf101, kuehne2011hmdb, kay2017kinetics, shahroudy2016ntu} originally from YouTube, movies or explicitly designed for recognition in household context.

Yet, a key  challenge  when applying energy expenditure models in practice arises  from transferring the learned concepts to new activity types.
To meet this requirement, \textsl{Vid2Burn} is equipped with a cross-category benchmark, where the caloric cost estimation models are evaluated against activity types not seen during training. 
In addition to potential mobile health applications, our dataset there fills the lack of benchmark studying concise recognition of body movement without relying on category-specific context biases.
From the computer vision perspective, the key technical challenges of our benchmark are (1) \textit{fine-grained} understanding of bodily movement and (2) \textit{generalization} to previously unseen types of activities. 
Extensive experiments with multiple state-of-the-art approaches for video- and body pose based action recognition demonstrate the difficulty of our task using modern video classification architectures, highlighting the need for further research.

\section{Related work}
\noindent\textbf{Activity recognition in videos.} 
Human activity recognition often operates on body poses~\cite{cheng2020skeleton,li2019actional,liu2020disentangling,shi2019skeleton,yan2018spatial} or learns representations end-to-end directly from the video data using Convolutional Neural Networks (CNNs)~\cite{carreira2017quo,hara2017learning,ji20123d,simonyan2014two,yue2015beyond}.
CNN-based approaches often deal with the temporal dimension via 3D convolution~\cite{qiu2017learning,tran2015learning,tran2018closer,xie2018rethinking,wang2018non} or follow the 2D+1D paradigm, chaining spatial 2D convolutions and subsequent 1D modules to aggregate the features temporally~\cite{feichtenhofer2017spatiotemporal,feichtenhofer2016convolutional,karpathy2014large,lin2019tsm,wang2016temporal,zhou2018temporal}.
Fueled by multiple publicly released large-scale activity recognition datasets collected from Youtube/Movies~\cite{kuehne2011hmdb,soomro2012ucf101,kay2017kinetics} or in home environments~\cite{shahroudy2016ntu}, the research of deep learning based activity recognition became a very active research field also explored in more targeted applications, \eg, in cooking~\cite{damen2018scaling,rohrbach2012database}, sports~\cite{parmar2019and}, robotics~\cite{karg2014human,rybok2011kit}, and automated driving~\cite{martin2019drive}-related tasks.
More specialized activity recognition research also addressed topics such as uncertainty of video classification models~\cite{roitberg2021uncertainty, tang2020uncertainty}.
However, all the approaches focus on \textit{categorization} into previously defined activity classes, while examining their feasibility for capturing complex physiological processes of the body, such as our calorie expenditure task, has been largely overlooked.

\begin{table}[t]
\centering
\scalebox{0.8}{
\begin{tabular}{lcccc} 
\toprule
\textbf{Datasets} & \multicolumn{1}{l}{\textbf{Vid2Burn-Diverse}} & \multicolumn{1}{l}{\textbf{Vid2Burn-ADL}} & \multicolumn{1}{l}{\textbf{Vid2Burn}} \\ 
\midrule
Video origin & \makecell{Youtube/\\movie datasets} & ADL datasets&  \makecell{Youtube/\\Movie/ADL}\\
\#Clips & 4260 & 5529 & 9789 \\
\#Activities & 33 & 39 & 72 \\
\makecell[l]{\#Train/test \\categories} & 27/6 & 33/6 & 60/12  \\
Unit & kcal/hour & kcal/hour & kcal/hour \\
Min & 64 & 153 & 64 \\
Max & 961 & 449 & 961 \\
Mean & 373 & 276 & 318 \\
\bottomrule
\end{tabular}}
\vskip-1ex
\caption{An overview of the main properties of  \textsl{Vid2Burn} and its two versions (including statistics of the caloric annotations).}
\vskip-3ex
\label{tab:comparison}
\end{table}

\noindent\textbf{Energy expenditure prediction.}
Visual estimation of calorie values  has been mainly investigated in food image analysis (\ie, tracking the amount of caloric intake)~\cite{naritomi2020caloriecaptorglass,ruede2021multitask_calorie,marin2019recipe1m}.
Energy \textit{expenditure} induced by physical activity is mostly studied from an egocentric perspective featuring data from wearable sensors, such as accelometors or heart rate monitors~\cite{albinali2010using,barut2020multitask,hedegaard2020prediction,kendall2019validity,meina2010combined,nakamura2017jointly,tao2018energy,o2020improving,xiao2020activity}, with a recent survey provided in~\cite{zhang2021energy}.
Only very few works address the  \textit{visual} predicted activity-related caloric expenditure~\cite{tao2018energy,nakamura2017jointly}.
The only dataset collected for energy expenditure prediction by visually observing the human~\cite{tao2018energy} features a highly simplistic evaluation setting (a single environment) and is comparably small in size, restricting the investigation of data-driven CNNs in this scenario (besides, the access to the collected database is restricted).
To the best of our knowledge, no previous work explored deep CNNs for estimation of caloric cost from human observation.
The research most similar to ours is presumably
the work of Nakamura \etal~\cite{nakamura2017jointly}, who collected an \textit{egocentric} video dataset for estimating the energy expenditure and explore CNN-based architectures for this task.
However, the research of~\cite{nakamura2017jointly} is significantly different from ours, as the cameras are mounted on the human, so that he/she is not observed.
Our dataset is created with the opposite perspective in mind, as we target caloric expenditure estimation \emph{from human video observations}.

\section{Vid2Burn: A Benchmark for Estimating Caloric Expenditure in Videos}

Given the growing demand for eHealth apps\footnote[3]{Grand View Research. mHealth Apps Market Size, Share \& Trends Analysis Report By Type (Fitness, Medical), By Region (North America, APAC, Europe, MEA, Latin America), And Segment Forecasts, 2021 - 2028. 2021. Available from: \url{www.grandviewresearch.com/industry-analysis/mhealth-app-market.}}, it is surprising that there is not a larger body of work on estimating physical intensity of activities in videos.
This might be due to the general focus of video classification research evolving mostly around activity \textit{categorization}~\cite{carreira2017quo,feichtenhofer2016convolutional,feichtenhofer2019slowfast, hara2017learning,chen2021deep}, while virtually all exercise intensity assessment datasets focus on \textit{wearable} sensors~\cite{albinali2010using,barut2020multitask,nakamura2017jointly} delivering, \eg, heart rate or accelerometer signals.
To promote the task of \textit{visually} estimating the hourly amount of kilocalories burned by the human during the current activity, we introduce the novel \textsl{Vid2Burn} dataset, featuring $>9K$ videos of $72$ different activity types with both caloric expenditure annotations on category- and sample-level.

\subsection{Dataset Collection}
 
\textsl{Vid2Burn} is an omni-source dataset developed with a diverse range of movements and settings in mind.
Our data collection procedure comprised the following steps.
We started by surveying the well-known available datasets for categorical activity classification, (\eg,~\cite{kay2017kinetics,shahroudy2016ntu,soomro2012ucf101,kuehne2011hmdb}).
Then, we identified categories which are not only accessible from these public datasets but also have great technically feasibility to infer caloric cost annotations.
The main sources of our dataset are UCF-101~\cite{soomro2012ucf101}, HMDB51~\cite{kuehne2011hmdb}, test set of Kinetics~\cite{kay2017kinetics} and NTU-RGBD~\cite{shahroudy2016ntu}.
We manually identified $72$ activity types for which the hourly caloric cost can be estimated based on the established physiological models, (\eg,~\cite{ainsworth20112011, tsou2015estimation,keytel2005prediction}).
Then, we estimated the labels for the energy expenditure based on these models on the category- and sample-level described in Section \ref{sec:annotation}.

\subsection{Dataset Structure}
The benefits of understanding caloric cost from videos extend to many applications, such as tracking of active exercise routines~\cite{jo2019there} or monitoring the daily physical activity level for elderly care~\cite{reeder2020older, jo2021elderly,awais2018physical}.
From the technical perspective, it is also useful to distinguish settings with higher and lower differences between the samples.
Lastly, while it is feasible to derive proper ground-truth  for coarse behaviours or situations with well-studied energy expenditure, (\eg, types of sports and exercises), many daily living activities do not fall into this category and should be addressed  with different techniques. Motivated by this, we group the content of \textsl{Vid2Burn} in two subsets: \textsl{Vid2Burn\textsubscript{Diverse}} and \textsl{Vid2Burn\textsubscript{ADL}}. Table \ref{tab:comparison} gives an overview of \textsl{Vid2Burn} and both its variants.

\textbf{\textsl{Vid2Burn-Diverse} }is collected from Youtube- and movie-based sources~\cite{soomro2012ucf101, kuehne2011hmdb, kay2017kinetics} and therefore features a highly uncontrolled environment (camera movement, diverse inside/outside backgrounds).
Since we focused on activities with well-studied energy expenditure models, a large portion of behaviours are related to sports, (\eg, \textsl{PushUps}). 
However, the database also covers certain everyday activities, such as \textsl{walking}, \textsl{standing} or \textsl{eating}.
The distribution of different activity types is summarized in Figure~\ref{fig:diverse}. 
On average, the dataset features $129$ video clips per category using category labels inherited from the original sources, with \textsl{walking} being unsurprisingly the most common behaviour for $548$ videos while  \textsl{stretching} and \textsl{shopping} are the least frequent ones for $47$ videos.

\textbf{\textsl{Vid2Burn-ADL}} on the other hand targets Activities of Daily Living (ADL) and might be used for physical workload tracking in smart homes. 
The activity types and video examples are derived from the public NTU-RGBD~\cite{shahroudy2016ntu} dataset for ADL classification and, compared to \textsl{Vid2Burn\textsubscript{Diverse}}, this dataset contains activities of rather lower physical intensities (\eg, \textsl{pickup}, \textsl{take off jacket}, \textsl{read}, \textsl{drink water}).
The environment of \textsl{Vid2Burn\textsubscript{ADL}} is much more controlled and the differences between the individual samples are at smaller scale.
In other words,\textsl{Vid2Burn\textsubscript{ADL}} can be regarded as a much more fine-grained benchmark for caloric cost regression. 
In contrast to \textsl{Vid2Burn\textsubscript{Diverse}}, the categories of \textsl{Vid2Burn\textsubscript{ADL}} are rather well-balanced and the number of examples per activity type is $142$ on average (detailed frequency statistics  provided in the supplementary).

\begin{figure}[t]
  \centering
  \includegraphics[width=\linewidth, angle =0]{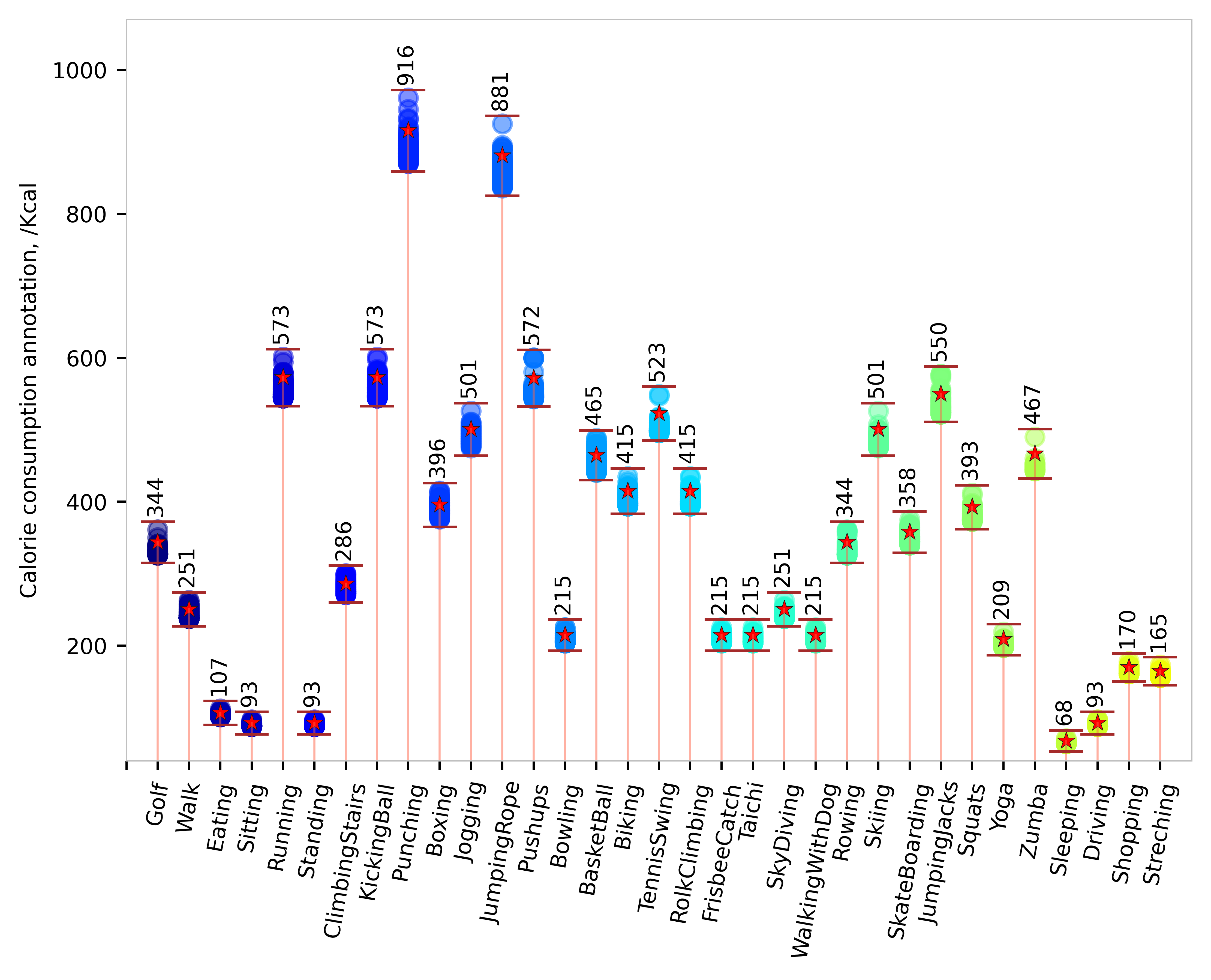} 
\vskip-1ex
\caption{
Statistical analysis of the sample-wise caloric cost values in \textsl{Vid2Burn\textsubscript{Diverse}}.
Red stars indicate the category-wise calorie consumption annotation, while the standard deviation boundaries result from sample-wise corrections based on the  body movement.
}
\vskip-3ex
\label{fig:analysis}
\end{figure}

\subsection{Caloric Expenditure Annotations}
\label{sec:annotation}

To adequately represent activity-induced caloric expenditure, we conducted a literature review on this physiological process~\cite{pinheiro2011energy,ainsworth20112011,withers2006self,tsou2015estimation,meina2010combined}.
Tracking the heat expended by nutrients oxidation (\ie, monitoring oxygen intake and carbon dioxide production) is considered the most accurate way of estimating energy expenditure~\cite{pinheiro2011energy}.
While this method is invasive and not practical for large-scale use, a multitude of topical studies conduct and publish such measurements for specific groups of activities, which is often summarized by meta-reviews in the form of compendiums~\cite{ainsworth20112011}. 
Such catalogues provide energy expenditure values for specific activities that are often available online as look-up tables.
A common way of estimating caloric cost, which is also leveraged by us, is deriving it from the heart-rate with validated physiological models~\cite{meina2010combined,keytel2005prediction}. Our annotation scheme leverages three methods to estimate hourly energy expenditure: established medical compendiums~\cite{ainsworth20112011}, heart-rate based measurements~\cite{keytel2005prediction} as well as adjustments based on the captured body movement~\cite{tsou2015estimation}. 

Next, we describe the three different ways for obtaining caloric cost ground truth (Section ~\ref{sec:annotation_methods}) and explain how we leveraged them to annotate the \textsl{Vid2Burn\textsubscript{Diverse}} (Section~\ref{sec:annotation_diverse}) and the \textsl{Vid2Burn\textsubscript{ADL}} (Section~\ref{sec:annotation_adl}) datasets.

\subsubsection{Annotation Methods}
\label{sec:annotation_methods}

Our derived annotation scheme leverages three types of sources: (1) current activity category, (2) intensity of the skeleton movement, as well as, (3) heart rate measurements obtained for a subset of activities (household activities) in a complementary study.

\mypar{Caloric cost values from published compendiums}
First, we leverage activity-specific metabolic rate values from published compendiums~\cite{ainsworth20112011} often summarized as look-up tables available on the web.\footnote{\url{https://captaincalculator.com/health/calorie/}}
For simplicity, we assume the body weight as $150~lb$, since this is also the average body weight of subjects captured in our heart rate measurement study.
Examples of the category-wise caloric expenditure annotations are marked as stars in Figure \ref{fig:analysis} for the\textsl{Vid2Burn\textsubscript{Diverse}} dataset.

\mypar{Heart-rate based annotations}
\textsl{Vid2Burn\textsubscript{ADL}} focuses on daily living activities, which naturally exhibit lower intensity of movement.
 The differences are at a much smaller scale compared to \textsl{Vid2Burn\textsubscript{Diverse}} and
the average expected energy expenditure has not been well-studied for many such concise types of physical activity. 
However, due to the more restricted nature of the environment, the NTU-RGBD setting is easy to reproduce. 
In such cases, we  recreate the environment of $39$ activities of \textsl{Vid2Burn\textsubscript{ADL}} and estimate their average caloric cost based on \textit{heart rate measurements} captured in a study with four volunteer participants.
Four people, one female and three males, participated in the data collection (1 female, 3 male, $27.75$ years old on average, average weight 150 lb).
The participation in our study was voluntary, and the subjects were instructed about the scope and purpose of the data collection and have given their written consent according to the requirements of our institution.

The heart rate of all participants was recorded using a wrist band activity tracker.\footnote{\url{https://mi.com/global/mi-smart-band-4/specs/}}
The subjects were asked to execute $39$ activities of the \textsl{Vid2Burn\textsubscript{ADL}} with a resting period in between to ensure the heart rate recovery.
More information about the study setup is provided in the supplementary.

Given the measured heart rate, we compute the caloric cost of the activity in accordance to~\cite{keytel2005prediction} as,
\begin{equation}
\begin{split}
    &Cal_M= 60\times T\times((-55.0969)+(0.6309\times HR)+\\
    &(0.1988\times W) + (0.2017 \times A))/4.184,
\end{split}
\end{equation}
\begin{equation}
\begin{split}
    &Cal_F= 60\times T\times((-20.4022)+(0.4472\times HR)-\\
    &(0.1263\times W) + (0.074 \times A))/4.184,
\end{split}
\end{equation}
where $Cal_M/Cal_F$ indicates the hourly caloric  expenditure for male/female, $HR$ is heart rate, $W$ is the body weight,  $A$ indicates participant's age and $T$ is the time length in hour.

\mypar{Body movement- based annotations}
Next, we approximate the caloric cost induced by the movement by leveraging the model of Tsou~\etal~\cite{tsou2015estimation}.
People can engage in the same type of activities in different ways and, since the amount of calories burned is directly linked to the amount/types of active muscles and the intensity, more active bodily movements lead to higher caloric cost.
Tsou~\etal~\cite{tsou2015estimation} formalizes and validates a model corresponding based on the movement of eight body regions $r$. 
We estimate the skeleton movement using \emph{AlphaPose}~\cite{fang2017rmpe,li2018crowdpose,xiu2018poseflow} for \textsl{Vid2Burn\textsubscript{Diverse}}, while for \textsl{Vid2Burn\textsubscript{ADL}} we use the skeleton data provided by the authors of the original datasets~\cite{montoye2019accuracy, shahroudy2016ntu}.
Following~\cite{tsou2015estimation}, we group skeleton joints into eight regions-of-interest to approximate the energy consumption as:
\begin{equation}
E_{body} = \sum_{t=1}^{F}\sum_{r=1}^{8}\omega_{r}\lambda ^2\left|\frac{1}{2}M_r \Delta x_{rt}^2+ \frac{1}{2}M_r\Delta y_{rt}^2+\frac{1}{2}M_r\Delta z_{rt}^2\right|
\label{eq:energy}
\end{equation}
where $\Delta x_{rt}, \Delta y_{rt}, \Delta z_{rt}$ indicate the position difference of the body region between frame $t$ and $t-1$ (the average position of all body joints inside same region), $F$ indicates the frame number and $\lambda$ is the frame frequency, $M_r$ indicates the mass of $r$th body region.
The final calorie consumption is obtained via multiplication with $0.239 (Cal/J) \times F\times \lambda \times3600 (h/s)$ in per hour-wise expression. 
The weighting factors $\omega_i$ of the different body regions are obtained from~\cite{tsou2015estimation}.
The main purpose of the body-pose based caloric cost estimation is to enable more concise annotations at \textit{sample-level}, since the same activity can be executed with different intensities. 

\subsubsection{\textsl{Vid2Burn-Diverse} annotations}
\label{sec:annotation_diverse}

One strategy behind the design of \textsl{Vid2Burn\textsubscript{Diverse}} was to select behaviour types for which the average caloric costs have been well-studied and easily accessible~\cite{ainsworth20112011} (for that reason, sports-related videos constitute a  significant portion of \textsl{Vid2Burn\textsubscript{Diverse}}). 
The main source for the category-level values is therefore derived from the published average category-specific  
which have been well-studied and easily accessible in this regard (on category level).
We then correct the estimations of the individual videos based on the previously explained body-movement model~\cite{tsou2015estimation}, resulting in more concise sample-level annotations.
\vspace{-0.3cm}
\subsubsection{\textsl{Vid2Burn-ADL} annotations}
\label{sec:annotation_adl}
Energy expenditure is not well-studied for many of the more concise daily living situations in \textsl{Vid2Burn\textsubscript{ADL}}.
We therefore take a detour by conducting a study with participants' heart rate recorded during the $39$ target activities (as described in the heart rate paragraph of Section \ref{sec:annotation_methods}).
For each activity type, we estimate (1) the average heart rate-based caloric cost value obtained from our study, (2) the average values based on the skeleton movement captured in the \textsl{Vid2Burn\textsubscript{ADL}} videos and (3) if available, values established from published medical compendiums~\cite{ainsworth20112011}.
Note, that while for \textsl{Vid2Burn\textsubscript{Diverse}} such estimations from published compendiums represent the main ground truth source, which are only available for $4$ out of $39$ \textsl{Vid2Burn\textsubscript{ADL}} activities. 
The final category-level annotations are then computed as the average of the estimations delivered by the two/three aforementioned methods. 
Similar to \textsl{Vid2Burn\textsubscript{Diverse}}, we then derive more accurate sample-level annotations through previously described skeleton-based corrections.

\subsection{Same- and intra-activity splits}
\label{sec:splits}
Since we specifically aim to rate generalization of the calorie estimation models for new activity types, we construct two testing scenarios covering (1) \textit{Known activity types} evaluation, with videos covering the same behaviours as the training set and (2) \textit{unknown activity types} evaluation where the train and test samples are drawn from different activity types. 
We randomly select $27$ (\textsl{Vid2Burn\textsubscript{Diverse}}) and $33$ (\textsl{Vid2Burn\textsubscript{ADL}}) activity types for the training set, while for both benchmark versions, the $6$ remaining categories are used for evaluation.
Next, the data of the $27/33$ training activity types is further split into training/testing (with ratio $7:3$) for the same-category evaluation.
Note that the category annotations used for constructing the splits are inherited from the source datasets.
Overall, our dataset comprises $2782/3243$ videos for training, $1192/1390$ samples for the validation on the same activity type and $286/896$ samples for new-activity-type evaluation for the \textsl{Vid2Burn\textsubscript{Diverse}} and \textsl{Vid2Burn\textsubscript{ADL}} databases, respectively.

\section{Activity Recognition Models in the Context of Caloric Cost Estimation}

\subsection{Learning  continuous caloric values}
\label{sec:smoothing}

Given a video input, our goal is to  infer hourly energy cost of the activity in which the depicted human is  involved.
Note, that we target the \textit{intensity} of the bodily activity and not its duration, \ie, our goal is to infer kilocalories burnt \textit{per hour}. Since our targets are \textit{continuous} caloric values our task naturally suits regression-based losses, such as the Euclidean L2 loss.
However, we observed that regression optimisation converges to a constant value in our case (a similar effect has been reported before in multimodal problems, \eg, in~\cite{zhang2016colorful}).
We therefore address this problem as multinomial classification with additional label softening.
Similar to~\cite{tang2020uncertainty}, we binarize each caloric value annotation $l$ with resolution of $1$ kcal inside a range $n \in [0, N]$, where $N$ is set to $1000$ kcal.
To keep certain regression properties, such as penalizing predictions which fall closer to the ground-truth bin less, we soften the labels through Gaussian distribution with a given standard deviation (STD) denoted as $\sigma$.
Then, for each ground truth annotation $l$, we obtain the softened label as a distribution over $N$ bins:
\begin{equation}
\begin{split}
    l_{s}[n] = \frac{1}{\sigma \sqrt{2\pi} }exp^{(-\frac{(n-l)^2}{2\sigma^2})},
\end{split}
\end{equation}
where $l_{s}$ indicates the soft label used for the supervision. 
We then use the the Kullback-Leibler (KL) divergence between the ground truth and predicted distributions:
\begin{equation}
\begin{split}
    KL_{loss} (y, l_s) = \frac{1}{N}\sum_{i=0}^{N-1}(l_s[n](logl_s[n])-y[n]),
\end{split}
\end{equation}
where $y$ indicates the predicted distribution with $n \in [0,N]$.

\subsection{Video representation backbones}

We adopt five modern video- and body pose-based  architectures developed for categorical human activity recognition as our video representation backbones.

\mypar{I3D} The Inflated 3D CNN (I3D) is a widely-used activity recognition backbone~\cite{carreira2017quo} and is a spatio-temporal version of the Inception-v1 network.
Weights transfer from pretrained 2D CNNs and its pretraining is achieved by repeating (``inflating'') the weights along the temporal axis.

\mypar{R3D} 
This 3D convolutional architecture~\cite{hara2017learning} with a remarkable depth of 101 layers (enabled through residual connections) chains multiple ResNeXt blocks, which are shallow three-layered networks leveraging group convolution. 

\mypar{R(2+1)D}
Unlike previous models, R(2+1)D~\cite{tran2018closer} ``mimics'' spatio-temporal convolution by factorizing
it into distinct 2D spatial and 1D temporal convolutions, yielding remarkable results despite these simpler operations.
This framework also leverages a residual architecture.

\mypar{SlowFast}
Our last CNN-based architecture is the SlowFast model of 
Feichtenhofer \etal\cite{feichtenhofer2019slowfast} which introduces two branches: a slow pathway and a fast pathway capturing cues from different temporal resolutions.

\mypar{ST-GCN}
In addition to the video-based models, we consider a popular architecture comprising a graph neural network operating on the estimated body poses~\cite{yan2018spatial}, which uses spatial- and temporal graph convolutional neural network to harvest human motion cues.
\begin{table}
\centering
\scalebox{0.65}{
\begin{tabular}{lllllllll} 
\toprule
\multicolumn{1}{c}{\multirow{2}{*}{\textbf{Method}}} &  & \multicolumn{3}{c}{\textbf{Known activity types }} &  & \multicolumn{3}{c}{\textbf{New activity types }} \\
\multicolumn{1}{c}{} &  & \textbf{\textcolor[rgb]{1,0.325,0.157}{MAE}}$\downarrow$ & \textbf{SPC}$\uparrow$ & \textbf{NLL }$\downarrow$ &  & \textbf{\textcolor[rgb]{1,0.325,0.157}{MAE} }$\downarrow$ & \textbf{SPC }$\uparrow$ & \textbf{NLL }$\downarrow$ \\ 
\midrule
Average &  & {\cellcolor[rgb]{0.753,0.753,0.753}}160.1 & - & - &  & {\cellcolor[rgb]{0.753,0.753,0.753}}220.7 & - & - \\
Random &  & {\cellcolor[rgb]{0.753,0.753,0.753}}306.4 & - & - &  & {\cellcolor[rgb]{0.753,0.753,0.753}}342.8 & - & - \\
ST-GCN &  & {\cellcolor[rgb]{0.753,0.753,0.753}}140.8 & 19.86 & 28.17 &  & {\cellcolor[rgb]{0.753,0.753,0.753}}285.1 & 4.41 & 14.01 \\ 
\midrule
SF-AVR &  & {\cellcolor[rgb]{0.753,0.753,0.753}}34.5 & 30.46 & 5.23 &  & {\cellcolor[rgb]{0.753,0.753,0.753}}159.5 & 8.20 & 11.50 \\
R3D-AVR &  & {\cellcolor[rgb]{0.753,0.753,0.753}}49.5 & 29.30 & 5.01 &  & {\cellcolor[rgb]{0.753,0.753,0.753}}154.3 & 8.56 & 14.58 \\
I3D-AVR &  & {\cellcolor[rgb]{0.753,0.753,0.753}}38.9 & 32.25 & 4.70 &  & {\cellcolor[rgb]{0.753,0.753,0.753}}228.7 & 7.61 & 16.32 \\
R(2+1)D-AVR &  & {\cellcolor[rgb]{0.753,0.753,0.753}}47.4 & 30.58 & 4.63 &  & {\cellcolor[rgb]{0.753,0.753,0.753}}246.9 & 10.52 & 14.26 \\
SF-LSTM &  & {\cellcolor[rgb]{0.753,0.753,0.753}}39.7 & 31.71 & 4.85 &  & {\cellcolor[rgb]{0.753,0.753,0.753}}227.3 & 10.45 & 17.41 \\
R3D-LSTM &  & {\cellcolor[rgb]{0.753,0.753,0.753}}82.7 & 28.93 & 5.23 &  & {\cellcolor[rgb]{0.753,0.753,0.753}}282.2 & 6.49 & \textbf{12.16} \\
I3D-LSTM &  & {\cellcolor[rgb]{0.753,0.753,0.753}}57.9 & 31.41 & 4.87 &  & {\cellcolor[rgb]{0.753,0.753,0.753}}251.9 & 8.77 & 12.26 \\
R(2+1)D-LSTM &  & {\cellcolor[rgb]{0.753,0.753,0.753}}53.5 & 31.26 & 4.82 &  & {\cellcolor[rgb]{0.753,0.753,0.753}}182.4 & 12.18 & 19.20 \\
\bottomrule
\end{tabular}}
\vskip-1ex
\caption{
Recognition results on the \textsl{Vid2Burn\textsubscript{Diverse}} benchmark in the \textit{category-wise} caloric annotations setting.
}
\label{tab:table1}
\vskip-1ex
\end{table}

\begin{table}
\vskip-1ex
\centering
\scalebox{0.59}{
\begin{tabular}{llllllllll} 
\toprule
\multicolumn{1}{c}{\multirow{3}{*}{\textbf{ Method }}} & \multicolumn{4}{c}{\textbf{Known activity types}} &  & \multicolumn{4}{c}{\textbf{New activity types}} \\
\multicolumn{1}{c}{} & \multicolumn{2}{c}{\textbf{Running }} & \multicolumn{2}{c}{\textbf{Climbing }} &  & \multicolumn{2}{c}{\textbf{Yoga }} & \multicolumn{2}{c}{\textbf{Shopping }} \\
\multicolumn{1}{c}{} & \multicolumn{1}{c}{\textbf{SPC }}$\uparrow$ & \multicolumn{1}{c}{\textbf{\textcolor[rgb]{1,0.369,0.259}{MAE} }$\downarrow$} & \multicolumn{1}{c}{\textbf{SPC }}$\uparrow$ & \multicolumn{1}{c}{\textbf{\textcolor[rgb]{1,0.369,0.259}{MAE} }$\downarrow$} &  & \multicolumn{1}{c}{\textbf{SPC }}$\uparrow$ & \multicolumn{1}{c}{\textbf{\textcolor[rgb]{1,0.369,0.259}{MAE} }$\downarrow$} & \multicolumn{1}{c}{\textbf{SPC }} $\uparrow$& \multicolumn{1}{c}{\textbf{\textcolor[rgb]{1,0.369,0.259}{MAE} }}$\downarrow$ \\ 
\midrule
ST-GCN & -13.51 & {\cellcolor[rgb]{0.753,0.753,0.753}}310.9 & 14.81 & {\cellcolor[rgb]{0.753,0.753,0.753}}225.6 &  & 19.65 & {\cellcolor[rgb]{0.753,0.753,0.753}}230.5 & -0.24 & {\cellcolor[rgb]{0.753,0.753,0.753}}322.3 \\ 
\midrule
SF-AVR & 11.76 & {\cellcolor[rgb]{0.753,0.753,0.753}}43.8 & 24.19 & {\cellcolor[rgb]{0.753,0.753,0.753}}57.7 &  & 20.18 & {\cellcolor[rgb]{0.753,0.753,0.753}}101.9 & 3.43 & {\cellcolor[rgb]{0.753,0.753,0.753}}174.4 \\
R3D-AVR & 6.14 & {\cellcolor[rgb]{0.753,0.753,0.753}}102.3 & 21.13 & {\cellcolor[rgb]{0.753,0.753,0.753}}135.5 &  & 17.82 & {\cellcolor[rgb]{0.753,0.753,0.753}}97.4 & 1.17 & {\cellcolor[rgb]{0.753,0.753,0.753}}108.5 \\
I3D-AVR & 7.69 & {\cellcolor[rgb]{0.753,0.753,0.753}}82.6 & 13.96 & {\cellcolor[rgb]{0.753,0.753,0.753}}81.8 &  & 13.10 & {\cellcolor[rgb]{0.753,0.753,0.753}}191.9 & 1.92 & {\cellcolor[rgb]{0.753,0.753,0.753}}190.5 \\
R(2+1)D-AVR & 12.88 & {\cellcolor[rgb]{0.753,0.753,0.753}}288.0 & 25.03 & {\cellcolor[rgb]{0.753,0.753,0.753}}74.0 &  & 15.20 & {\cellcolor[rgb]{0.753,0.753,0.753}}217.9 & 4.68 & {\cellcolor[rgb]{0.753,0.753,0.753}}271.1 \\
SF-LSTM & 12.59 & {\cellcolor[rgb]{0.753,0.753,0.753}}9.6 & 15.79 & {\cellcolor[rgb]{0.753,0.753,0.753}}86.7 &  & 19.49 & {\cellcolor[rgb]{0.753,0.753,0.753}}133.6 & 3.91 & {\cellcolor[rgb]{0.753,0.753,0.753}}266.8 \\
R3D-LSTM & 6.54 & {\cellcolor[rgb]{0.753,0.753,0.753}}262.1 & 8.82 & {\cellcolor[rgb]{0.753,0.753,0.753}}83.8 &  & 15.66 & {\cellcolor[rgb]{0.753,0.753,0.753}}223.2 & 0.02 & {\cellcolor[rgb]{0.753,0.753,0.753}}370.1 \\
I3D-LSTM & 6.09 & {\cellcolor[rgb]{0.753,0.753,0.753}}147.7 & 14.85 & {\cellcolor[rgb]{0.753,0.753,0.753}}99.9 &  & 12.04 & {\cellcolor[rgb]{0.753,0.753,0.753}}217.6 & 4.15 & {\cellcolor[rgb]{0.753,0.753,0.753}}280.0 \\
R(2+1)D-LSTM & 11.22 & {\cellcolor[rgb]{0.753,0.753,0.753}}37.1 & 10.47 & {\cellcolor[rgb]{0.753,0.753,0.753}}98.4 &  & 18.26 & {\cellcolor[rgb]{0.753,0.753,0.753}}181.8 & 8.53 & {\cellcolor[rgb]{0.753,0.753,0.753}}134.4 \\

\bottomrule
\end{tabular}}
\vskip-1ex
\caption{Prediction quality for individual activity types on \textsl{Vid2Burn\textsubscript{Diverse}} (for two known and two new categories).}
\label{tab:table2}
\vskip-5ex
\end{table}

\subsection{Temporal fusion}

Temporal windows captured by the above backbones vary between $16$ frames for I3D~\cite{carreira2017quo}, R(2+1)~\cite{tran2018closer} and R3D~\cite{tran2015learning} and $32$ frames for SlowFast~\cite{feichtenhofer2016convolutional} are considerably smaller than the durations of the video clips captured in \textsl{Vid2Burn}. 
Given an input video of length $T$ and a model $f_{\theta}(\cdot)$ which takes as input $F$ frames, we sequentially pass ${K}$ video snippets $\{\textbf{t}_1, \textbf{t}_2,...,\textbf{t}_K\}$ using sliding window with overlapping resulting in $K$ predictions $\{f_{\theta}(\textbf{t}_1), f_{\theta}(\textbf{t}_2),...,f_{\theta}(\textbf{t}_K)\}$.
We now consider two different strategies for fusing these results: (1) averaging of the output of the last fully connected layer and (2) learning to fuse the output with an additional LSTM network.
The first method employs simple average pooling of the final representation: $pred(\textbf{t}) = \frac{1}{K}\sum_{i=1}^{K} f_{\theta}(\textbf{t}_i)$.
Our second fusion strategy passes the last fully connected layer of our video representation backbone to an LSTM network with two layers with the number of neurons corresponding to the input size, trained together with the backbone model in an end-to-end fashion. 
Since an LSTM also produces sequential output, we also average the resultant sequence to obtain the final prediction.

\section{Experiments}

\subsection{Evaluation Protocols}

We adopt Mean Average Error (MAE) as our main evaluation metric and additionally report the Spearmann Rank Correlation (SPC),  and the Negative Logarithm Likelihood (NLL).
MAE is an intuitive metric reporting the mean disarray between prediction and ground truth in our target units, (\ie, kilocalories).
Note, that while SPC illustrates the association strength between the ground truth and the predictions (it is $+1$ if one is a perfect monotone function of the other and $-1$ if they are fully opposed), it should be interpreted with care, since it ignores scaling and shifting of the data.
In other words, SPC would not reflect if the number of kilocalories is constantly over/underestimated by similar amounts. It therefore should only be viewed as a complementary metric. Note, that we report SPC in \% for better readability (\ie, we multiply the result by $100$). Our experiments are carried out in two annotation settings: category- and sample-wise annotations (see Section \ref{sec:annotation} for details).
We view the sample-wise version as a more concise choice, but using static category labels is the protocol used in the past energy expenditure work from egocentric data~\cite{nakamura2017jointly} and is adopted for consistency.
As explained in Section~\ref{sec:splits}, we conduct the evaluations on both: behaviours present during training and new activity types.

\begin{table}[t]
\centering
\scalebox{0.76}{
\begin{tabular}{llllllll} 
\toprule
\multirow{2}{*}{\textbf{Method}} & \multicolumn{3}{c}{\textbf{Known activity types}} &  & \multicolumn{3}{c}{\textbf{New activity types}} \\
 & \textcolor[rgb]{1,0.369,0.259}{\textbf{MAE}}$\downarrow$ & \textbf{SPC}$\uparrow$ & \textbf{NLL}$\downarrow$ &  & \textcolor[rgb]{1,0.369,0.259}{\textbf{MAE}}$\downarrow$ & \textbf{SPC}$\uparrow$ & \textbf{NLL}$\downarrow$ \\ 
\midrule
Average & {\cellcolor[rgb]{0.753,0.753,0.753}}160.7 & - & - &  & {\cellcolor[rgb]{0.753,0.753,0.753}}220.7 & - & - \\
Random & {\cellcolor[rgb]{0.753,0.753,0.753}}308.2 & - & - &  & {\cellcolor[rgb]{0.753,0.753,0.753}}341.1 & - & - \\ 
\midrule
SF-AVR & {\cellcolor[rgb]{0.753,0.753,0.753}}57.9 & 49.41 & 6.08 &  & {\cellcolor[rgb]{0.753,0.753,0.753}}134.0 & 42.20 & 9.02 \\
R3D-AVR & {\cellcolor[rgb]{0.753,0.753,0.753}}140.6 & 42.13 & 7.01 &  & {\cellcolor[rgb]{0.753,0.753,0.753}}179.8 & 25.12 & 9.33 \\
I3D-AVR & {\cellcolor[rgb]{0.753,0.753,0.753}}128.8 & 44.18 & 6.73 &  & {\cellcolor[rgb]{0.753,0.753,0.753}}130.8 & 32.67 & 8.58 \\
R(2+1)D-AVR & {\cellcolor[rgb]{0.753,0.753,0.753}}98.9 & 52.21 & 6.53 &  & {\cellcolor[rgb]{0.753,0.753,0.753}}224.6 & 21.92 & 9.28 \\

\bottomrule
\end{tabular}}
\vskip-1ex
\caption{Recognition results on the \textsl{Vid2Burn\textsubscript{Diverse}} benchmark in the \textit{sample-wise} caloric annotations setting.}
\label{tab:table9}
\vskip-1ex
\end{table}

\begin{table}
\centering

\scalebox{0.65}{
\begin{tabular}{lcccc} 
\toprule
\multicolumn{1}{c}{\textbf{Method}} & \multicolumn{1}{c}{\textbf{\#Params}} & \multicolumn{1}{c}{\textbf{Inference time}} & \textbf{Same-MAE}$\downarrow$ & \textbf{New-MAE}$\downarrow$ \\ 
\midrule
SF-AVR & 34.87M & 102.44ms & 34.5 & 159.5 \\
R3D-AVR & 34.04M & 27.19ms & 49.5 & 154.3 \\
I3D-AVR & 12.75M & 65.22ms & 38.9 & 228.7 \\
R(2+1)D-AVR & 33.87M & 66.61ms & 47.4 & 246.9 \\
SF-LSTM & 51.66M & 102.73ms & 39.7 & 227.3 \\
R3D-LSTM & 50.82M & 27.30ms & 82.7 & 282.2 \\
I3D-LSTM & 29.54M & 65.75ms & 57.9 & 251.9 \\
R(2+1)D-LSTM & 50.67M & 66.82ms & 53.5 & 182.4 \\ 
\bottomrule
\end{tabular}}
\vskip-1ex
\caption{Nr. of parameters, inference time and MAE of different approaches (on\textsl{Vid2Burn\textsubscript{Diverse}}, Quadro-RTX 6000 graphic card).}
\label{tab:table7}
\vskip-1ex
\end{table}

\begin{figure}[t]
\vskip-1ex
\begin{center}
\includegraphics[width = .43\textwidth]{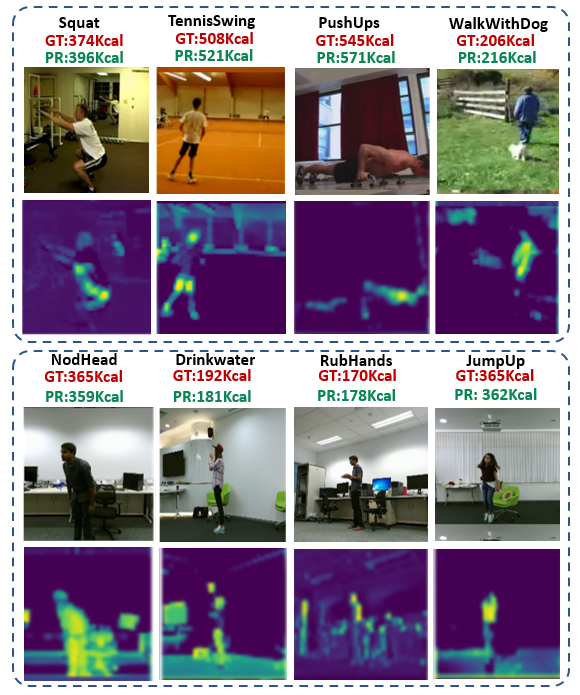}
\end{center}
\vskip-4ex
\caption{Qualitative results and visualization of the CNN activations for \textsl{Vid2Burn\textsubscript{Diverse}} (top) and \textsl{Vid2Burn\textsubscript{ADL}} (bottom).
We visualize the activations of the second convolutional layer (I3D).
PR and GT indicate the prediction and the ground truth.
}
\label{fig:vis}
\vskip-4ex
\end{figure}  

\subsection{Experiment Results}

All implemented models greatly outperform the random and average baselines in all
measures (Table \ref{tab:table1}), showing that our problem is feasible. 
 However, our experiments also underline the difficulty of estimating the energy cost if the model is evaluated in new previously unseen situations, motivating further research of models with deeper and more fine-grained understanding of physical activities.
 
Tables \ref{tab:table1} and \ref{tab:table9} illustrate the performance of the employed algorithms as well as the random and average baselines for \textsl{Vid2Burn\textsubscript{Diverse}} in the category- and sample-labelled settings together with different temporal fusion schemes, marked as \textsl{AVR} (for the average pooling fusion) and \textsl{LSTM}, with \textsl{AVR} consistently leading notably better results.
SlowFast (SF) with AVG consistently achieves the best recognition quality with only $34.5$ kcal/$20.1$ kcal MAE.
As expected, the results are lower in the more fine-grained sample-wise setting, since the backbones were initially developed for coarser categorization.
The task of caloric cost regression in previously unseen situations is much more difficult and the performance drops significantly. The MAE is $>4$ and $>2$ times higher in the category- and sample-wise settings separately for SF-AVG.
Interestingly, the best model in the case of \textsl{known activities} is usually not the top-performing approach for the \textsl{new} activity types, which is SF-AVG in case of the sample-wise evaluation of \textsl{Vid2Burn\textsubscript{Diverse}} with the MAE of $130.8$ kcal, but the gap to SF-AVG (MAE of $134$ kcal) is very small (Table \ref{tab:table9}).

Presumably due to a more restricted environment and smaller average caloric cost values, the models are more accurate on \textsl{Vid2Burn\textsubscript{ADL}} (Table \ref{tab:table10}). 
Consistently with the \textsl{Vid2Burn\textsubscript{Diverse}} results, SF-AVG yields the best recognition quality on known activities (MAE of $20.1$ kcal) and is the second best performing model on the new ones. 
Table \ref{tab:table10} also lists our results achieved with a regression loss (L2), marked as R3D-Reg. 
As explained in Section \ref{sec:smoothing}, we observe convergence to a constant value resulting in a very high MAE. 
Note that SPC cannot be computed in this case, since the output is a number while not a distribution.
We also consider the trade-off between the performance and the computational cost (Table \ref{tab:table7}).
\textsl{SF-AVG} offers a good balance between speed and accuracy, while the \textsl{I3D} backbone is a more lightweight model, but the MAE is higher.

\begin{table}[t]
\centering

\scalebox{0.65}{
\begin{tabular}{llllllll} 
\toprule
\multirow{2}{*}{\textbf{Method}} & \multicolumn{3}{c}{\textbf{Known activity types}} &  & \multicolumn{3}{c}{\textbf{New activity types}} \\
 & \textcolor[rgb]{1,0.369,0.259}{\textbf{MAE}} $\downarrow$& \textbf{SPC} $\uparrow$& \textbf{NLL}$\downarrow$ &  & \textcolor[rgb]{1,0.369,0.259}{\textbf{MAE}}$\downarrow$ & \textbf{SPC}$\uparrow$ & \textbf{NLL}$\downarrow$ \\ 
\midrule
R3D-Reg & {\cellcolor[rgb]{0.753,0.753,0.753}}281.7 & - & - &  & {\cellcolor[rgb]{0.753,0.753,0.753}}281.7 & - & - \\ 
Average & {\cellcolor[rgb]{0.753,0.753,0.753}}76.2 & - & - &  & {\cellcolor[rgb]{0.753,0.753,0.753}}59.4 & - & - \\ 
Random & {\cellcolor[rgb]{0.753,0.753,0.753}} 311.7& - & - &  & {\cellcolor[rgb]{0.753,0.753,0.753}}309.8 & - & - \\ 
\midrule
SF-AVR & {\cellcolor[rgb]{0.753,0.753,0.753}}20.1 & 68.61 & 5.57 &  & {\cellcolor[rgb]{0.753,0.753,0.753}}36.4 & 72.30 & 6.69 \\
R3D-AVR & {\cellcolor[rgb]{0.753,0.753,0.753}}35.1 & 76.91 & 6.01 &  & {\cellcolor[rgb]{0.753,0.753,0.753}}44.3 & 79.78 & 6.21 \\
I3D-AVR & {\cellcolor[rgb]{0.753,0.753,0.753}}22.9 & 72.97 & 5.66 &  & {\cellcolor[rgb]{0.753,0.753,0.753}}39.6 & 70.45 & 6.38 \\
R(2+1)D-AVR & {\cellcolor[rgb]{0.753,0.753,0.753}}26.2 & 71.31 & 5.99 &  & {\cellcolor[rgb]{0.753,0.753,0.753}}29.5 & 75.30 & 6.28 \\
SF-LSTM & {\cellcolor[rgb]{0.753,0.753,0.753}}36.9 & 77.19 & 6.05 &  & {\cellcolor[rgb]{0.753,0.753,0.753}}47.1 & 80.43 & 6.24 \\
R3D-LSTM & {\cellcolor[rgb]{0.753,0.753,0.753}}92.0 & 44.46 & 6.67 &  & {\cellcolor[rgb]{0.753,0.753,0.753}}111.7 & 53.30 & 6.63 \\
I3D-LSTM & {\cellcolor[rgb]{0.753,0.753,0.753}}101.4 & 42.42 & 6.68 &  & {\cellcolor[rgb]{0.753,0.753,0.753}}68.1 & 51.11 & 6.66 \\
R(2+1)D-LSTM & {\cellcolor[rgb]{0.753,0.753,0.753}}88.8 & 44.47 & 6.59 &  & {\cellcolor[rgb]{0.753,0.753,0.753}}72.9 & 54.97 & 6.52 \\

\bottomrule
\end{tabular}}
\vskip-1ex
\caption{Recognition results on the \textsl{Vid2Burn\textsubscript{ADL}} benchmark in the sample-wise caloric annotations setting.
}
\label{tab:table10}
\vskip-4.5ex
\end{table}

We further look at the recognition for the individual activity types: two known (\textsl{running}, \textsl{climbing}) and two unknown (\textsl{yoga}, \textsl{shopping}) behaviours as presented in Table \ref{tab:table2} (more results are provided in the supplementary).
The recognition quality varies greatly depending on whether the activity is familiar.
For example, SF-AVG is only off by $43.8$ kcal for \textsl{running}, but the MAE is $174.4$ kcal for \textsl{shopping}.
Finally, in Figure \ref{fig:vis} we showcase multiple qualitative results by visualizing the activation region of a CNN (I3D backbone) with multiple examples of
representative qualitative results for the \textsl{Vid2Burn\textsubscript{Diverse}} (top) and \textsl{Vid2Burn\textsubscript{ADL}} (bottom) datasets.
Additionally to the predicted caloric value and the ground truth, we visualize the activation regions of an intermediate CNN layer (in this case we choose the second convolutional layer). 
It is evident, that the largest focus is put on the body region activated during movement which we view as a positive property, since the energy expenditure is a direct result of the muscle movement~\cite{caspersen1985physical}.
However, in several cases the objects (\eg, the television) are clearly highlighted despite no direct interaction, indicating category-specific biases which are presumably the leading cause of mistakes in cross-category settings.

\mypar{Implementation Details}
Our models are trained with ADAM~\cite{kingma2014adam} using a weight decay of $1e^{-5}$, a batch size of $4$, a learning rate of $1e^{-4}$ for $40$ epochs, the model weights from Kinetics~\cite{kay2017kinetics} and a Quadro-RTX 6000 graphic card (parameter numbers and inference times reported in Table \ref{tab:table7}).
For binarization of the continuous label space, the maximum calorie prediction limit is set to $1000$ kcal for \textsl{Vid2Burn\textsubscript{Diverse}} and $500$ kcal for \textsl{Vid2Burn\textsubscript{ADL}} with resolution of $1$ kcal.
A more detailed description of the parameter settings is provided in the supplementary.
\vspace{-0.1cm}

\section{Conclusion}
We introduced the novel \textsl{Vid2Burn} benchmark for estimating the amount of calories burned during physical activities by visually observing the human.
Through our experiments, we found that the generalization ability of modern video classification CNNs is limited in this challenging task and we will keep tackling this issue in the future work.
\textsl{Vid2Burn} will be publicly released, opening new perspectives on specific challenges in human activity analysis, such as fine-grained understanding of bodily movement and generalization to new physical activity types, since our benchmark specifically evaluates the quality of energy expenditure estimation in \textit{new} situations.
We hope to foster research of human understanding models which are able to capture cues of the underlying physiological processes, (\eg, active muscles and their intensity) instead of learning rigid category-specific biases seen during training.

\vspace{0.1cm}
\mypar{Broader Impact and Limitations}
This work targets energy expenditure estimation from videos.
The benefits of such methods extend to multiple applications, such as supporting  healthy lifestyle, \eg, by tracking exercise routines~\cite{jo2019there} or monitoring the daily physical activity level for elderly care~\cite{reeder2020older}.
However, both annotations in our dataset and the results inferred by our models are \textit{approximations} and not exact measurements, which should be carefully used in medical care applications, as they are simplified by assuming the body weight of $150~lb$, while gender, height and age  are not taken into account.
Moreover, our data-driven algorithms may learn shortcuts and biases present in the data potentially resulting in a false sense of security.


{\small
\bibliographystyle{ieee_fullname}
\bibliography{egbib}
}

\setcounter{section}{0}
\renewcommand\thesection{\Alph{section}}

\section{Limitations}
In addition to the summary of limitations mentioned at the end of our main paper, more details about the limitations of our approaches and proposed benchmarks will be given in this section. 
This work targets estimation of energy expenditure from videos. 
The benefits of such methods extend to multiple applications, such as supporting active and healthy lifestyle, \eg, by tracking exercise routines~\cite{jo2019there} or monitoring the daily physical activity level for elderly care~\cite{reeder2020older, jo2021elderly,awais2018physical}.
However, our work is not without limitations.
Energy expenditure is a complex physiological process~\cite{caspersen1985physical}, and while bodily movement, (\ie active muscled and intensities) are its primary drivers, there is a variety of the contributing factors, such as age, gender, weight and personal metabolic rate. Many of these factors are not considered in our work.
For example, for simplicity, we derive energy annotations from medical compendiums assuming the weight of $150~lb$ (our study with the heart-rate based ground truth estimation is an exception, where age/gender/weight were taken into account).
The ground truth values of our dataset are therefore only \textit{approximate estimates}.
Furthermore, as with most data-driven algorithms, our models may learn shortcuts and biases presenting in the data (in our cases oftentimes category- and context-related biases), which may cause a false sense of security.
Direct caloriometry~\cite{pinheiro2011energy} or heart rate-based estimation~\cite{ainsworth20112011} are more accurate ways to estimate caloric cost than visual models.
\begin{figure}[t]
\begin{subfigure}[b]{0.45\textwidth}
  \includegraphics[width=\linewidth]{figure/Video2Burn_2.png}
 \caption{\textsl{Vid2Burn\textsubscript{Diverse}}}
  \label{fig:analysis1}
\end{subfigure}
\hfill
\begin{subfigure}[b]{0.45\textwidth}
  \includegraphics[width=\linewidth]{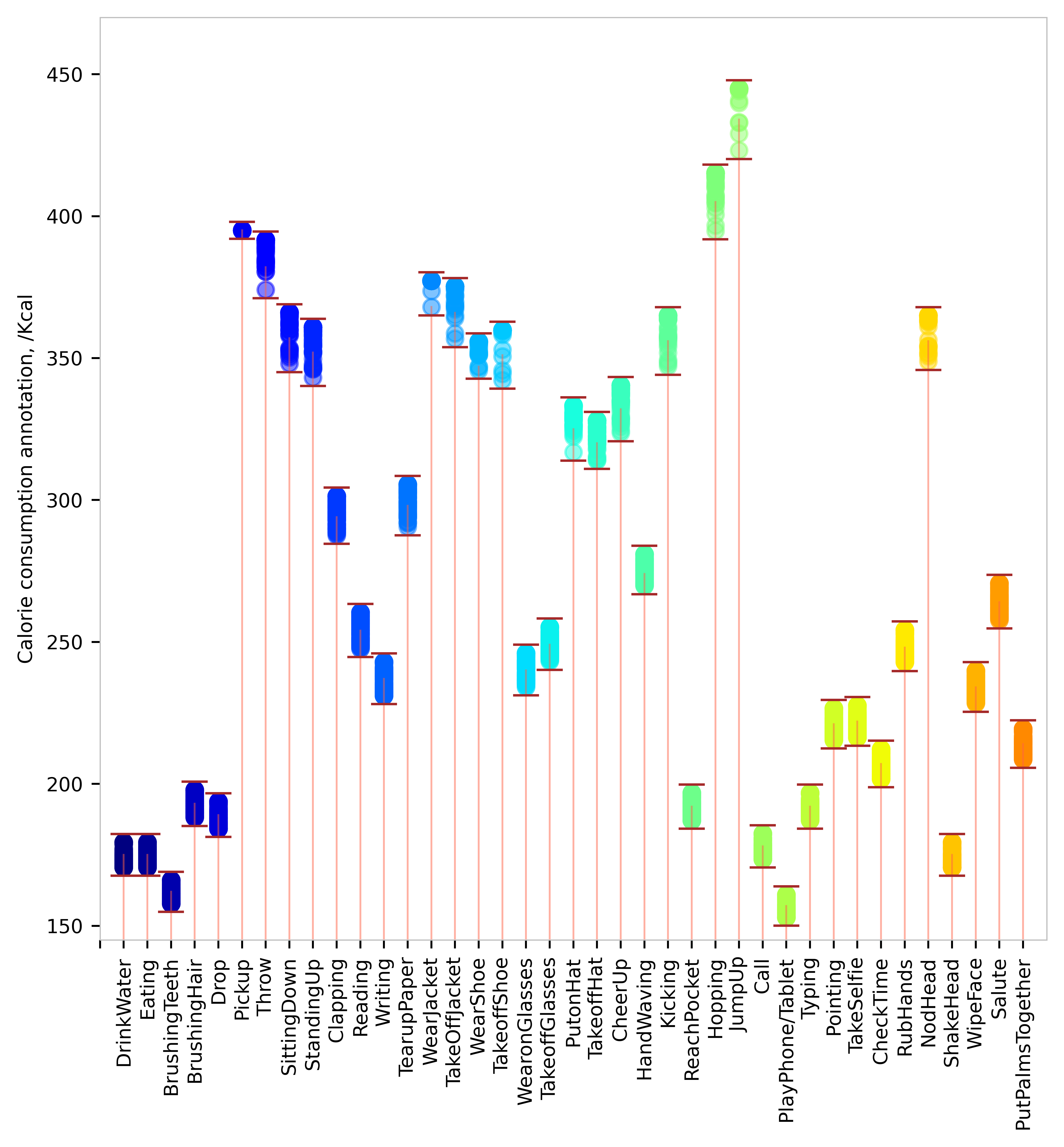}  
  \caption{\textsl{{Vid2Burn\textsubscript{ADL}}}}
  \label{fig:analysis}
\end{subfigure}
\caption{An overview of the calorie consumption annotation for \textsl{Vid2Burn} dataset. The boundaries of the fluctuations for each category is marked as brown lines, which define the sample-wise calorie consumption annotation based on the intensity of skeleton movement.}
\label{fig:analyses_all}
\end{figure}
\begin{figure*}[t]
\begin{center}
\includegraphics[width = 1\textwidth]{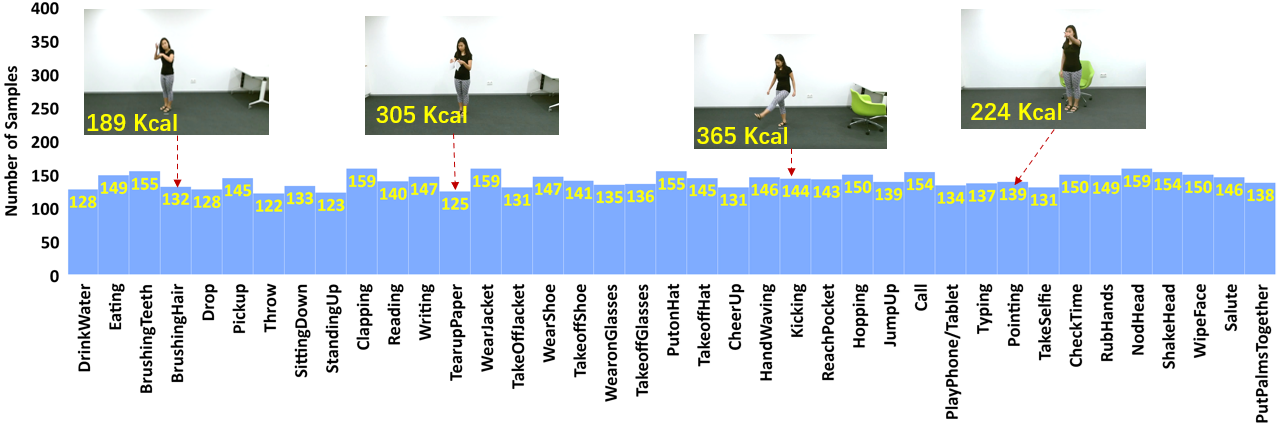}
\end{center}
\caption{An overview of the dataset structure for \textsl{Video2Burn\textsubscript{ADL}} dataset. The number in each histogram indicates the number of samples for the corresponding category and the number on each cluster of image represents the category-wise calorie consumption ground truth.}
\label{fig:adl}
\end{figure*} 
\vspace{-0.2cm}
\section{Broader impact}
\vspace{-0.2cm}
Our work introduces two video-based calorie consumption estimation benchmarks -- \textsl{Vid2Burn\textsubscript{Diverse}} and \textsl{Vid2Burn\textsubscript{ADL}}, together with several deep learning-based baselines targeting at end-to-end calorie consumption estimation. A wide range applications for health monitoring and human physical movement level prediction will directly benefit from this work. Moreover, since our work also tackle the generalization issue through evaluating the calorie estimation performance on the unseen activity types which can simulate the scenario for facing with out-of-distribution samples. The baselines leveraged in our work show a certain performance difference between the evaluations of known and unknown action types, indicating that offensive predictions, biased content and possible misclassifications can result in false sense of security while it still points out a valuable future research direction to us for further investigation. To allow future work constructed based on our benchmarks and baselines, we will make our code, models, and data publicly available.
\vspace{-0.2cm}
\section{License of existing asserts}
Since we use multiple public datasets and online resources to form the video dataset and annotation set, we have carefully cited the related works for these leveraged datasets and marked the website link of the online resources in the corresponding footnote in our paper.
\vspace{-0.2cm}
\section{Human subjects data collection clarification}
In \textsl{Vid2Burn\textsubscript{ADL}} dataset, we collect the heart rate, body weight and age data from 4 subjects to improve the accuracy of our calorie consumption annotation. The collected data is only leveraged to generate the global calorie consumption annotation which is highly aggregated and can not directly identify a specific person. The data and annotation are all anonymous. During the data collection procedure, each subject is well instructed to collect the heart rate data through wrist band (MIBAND 4) which can't bring any negative impact to the human body. From the dataset \textsl{Vid2Burn} which will be published soon, no person data is involved since all the data are highly aggregated. All participants are voluntary and signed a data collection agreement. We did not place the signed form for voluntary data collection in the supplementary materials in order to ensure anonymous submission.
\vspace{-0.2cm}
\section{Supplementary for \emph{Vid2Burn} dataset}
\subsection{Comparison between \emph{Vid2Burn} and other human energy expenditure datasets}
\begin{figure*}[t]
  \includegraphics[width=\linewidth]{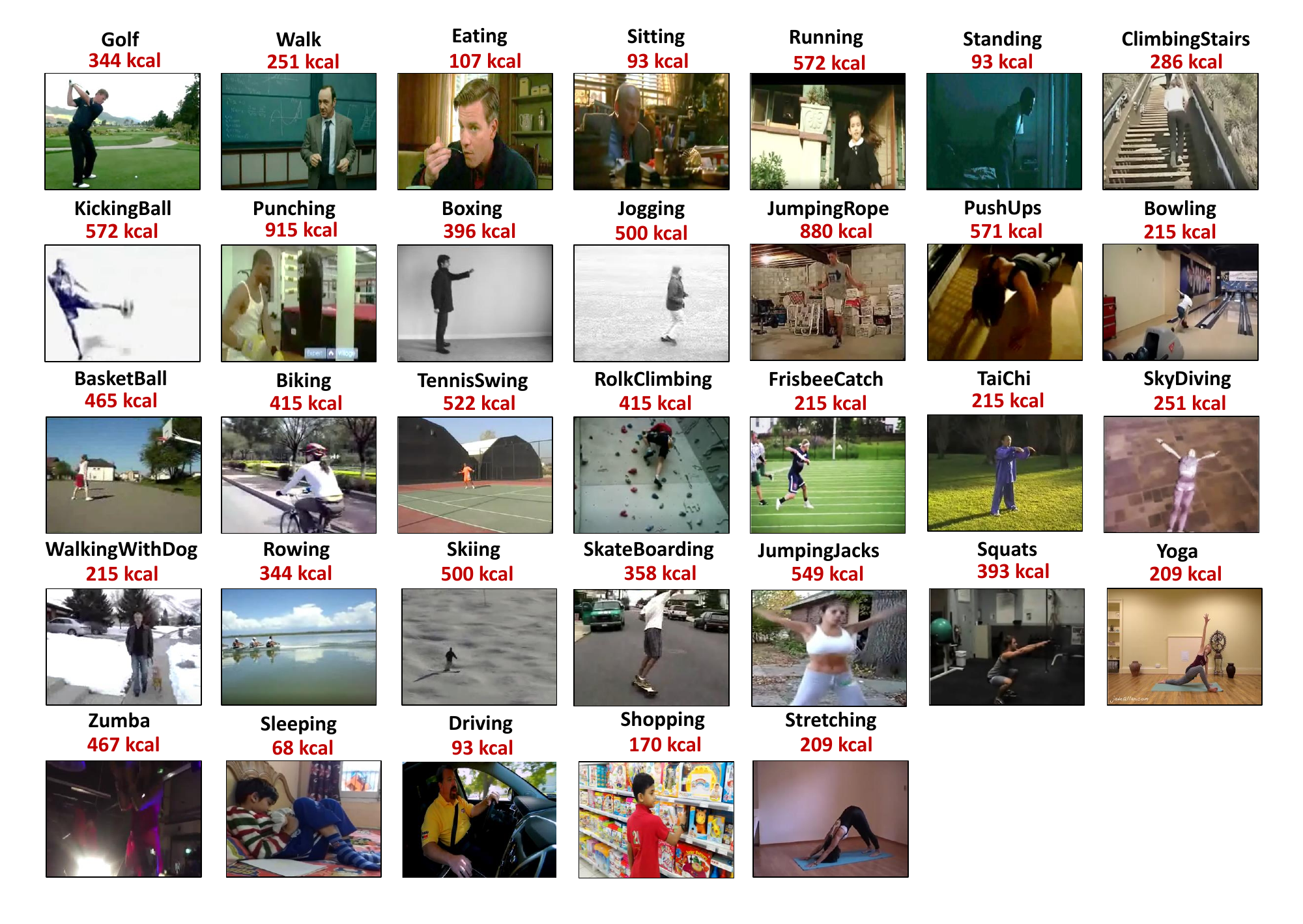}  
\caption{An overview of the calorie consumption annotation for \textsl{Vid2Burn\textsubscript{Diverse}} dataset for all $33$ leveraged action types (sample-wise annotation). We mark the corresponding calorie consumption annotation under each category name for each selected sample.}
\label{fig:diverse_all}
\end{figure*}

\begin{figure*}[t]
  \includegraphics[width=\linewidth]{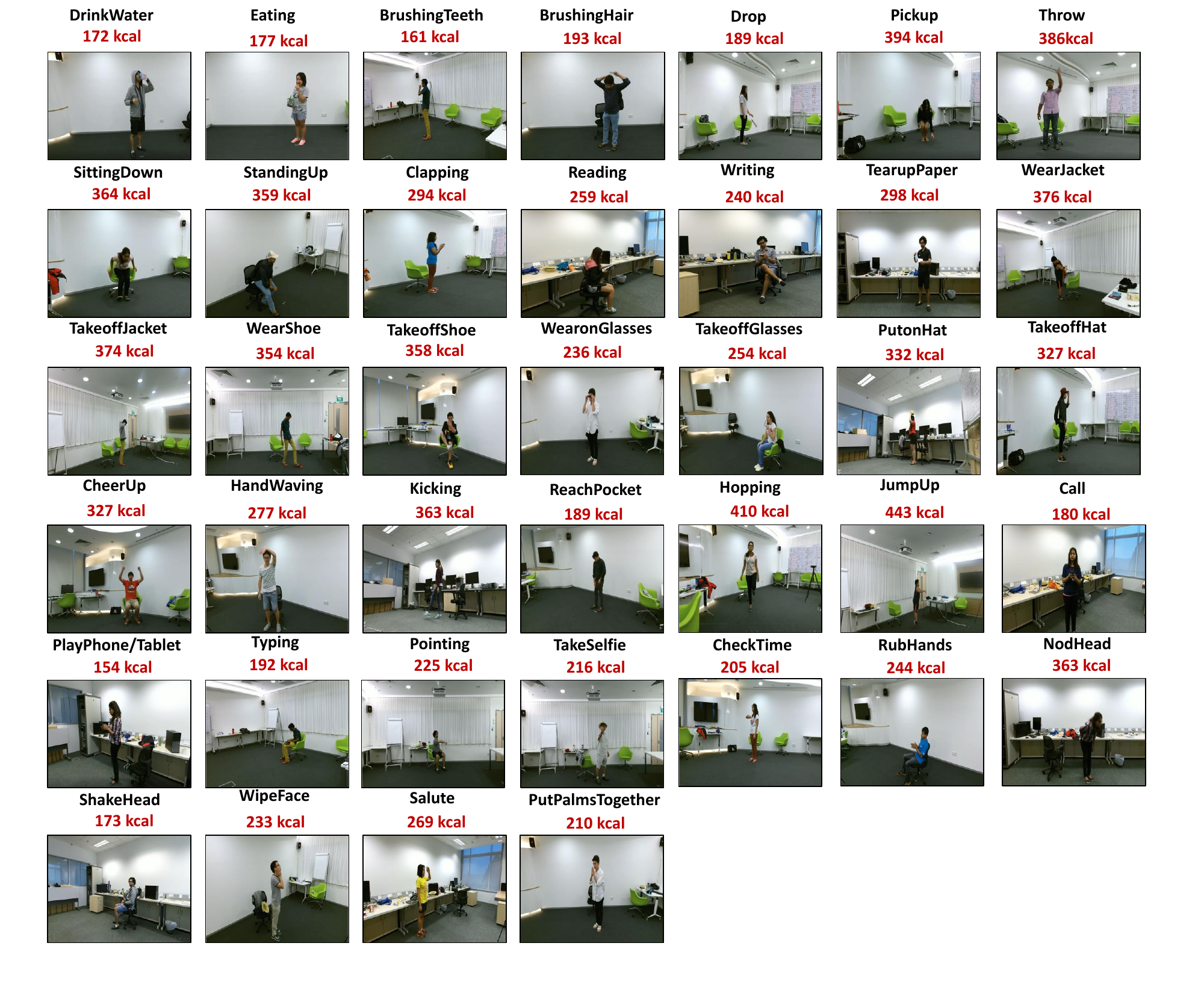}  
\vskip-3ex
\caption{An overview of the calorie consumption annotation for \textsl{Vid2Burn\textsubscript{ADL}} dataset for all $39$ leveraged action types (sample-wise annotation). We mark the corresponding calorie consumption annotation under each category name for the selected sample.}
\label{fig:adl_all}
\end{figure*}

In order to further clarify the strengthens of our proposed benchmarks, we make a comparison between the proposed two benchmarks, -- \textsl{Vid2Burn\textsubscript{Diverse}} and \textsl{Vid2Burn\textsubscript{ADL}}, and the other two existed video-based benchmarks which are Stanford-ECM~\cite{nakamura2017jointly} and Sphere~\cite{tao2018energy}. The camera setting for our proposed benchmarks and Sphere are all fixed-position while Stanford-ECM leveraged egocentric perspective requiring the camera to be mounted on a wearable device which limits the comfort of the user and requires contact, if application has been taken into consideration. Concerning the action numbers, our \textsl{Vid2Burn} contains in total $72$ kinds of activities which contains simultaneously high- and low-intensity activities, together with >9K video clips which is much larger compared with the other two datasets offering more possibility to achieve deep learning based end-to-end calorie consumption estimation. In addition, our benchmarks provide sample-wise calorie consumption annotation which is more precise compared with the other datasets that only provide category-level human energy expenditure annotations. We also provide the description of the two proposed benchmarks in Figures \ref{fig:diverse_all} and \ref{fig:adl_all} to show part of the label-sample pair with sample-wise calorie consumption annotation for each benchmark. There are $33$ and $39$ label-sample pairs for \textsl{Vid2Burn\textsubscript{Diverse}} and \textsl{Vid2Burn\textsubscript{ADL}} separately.
\subsection{Supplementary for Vid2Burn-ADL dataset}

First, a detailed introduction of sample numbers under each activity type, indicated by the number of samples on each histogram, and the corresponding category-wise annotation, denoted by the number on each image, are introduced in Fig.~\ref{fig:adl}. The sample numbers for different actions show a balanced distribution with minimum sample number as $122$ and maximum sample number as $159$.
Second, the statistic analysis for \textsl{Vid2Burn\textsubscript{ADL}} dataset is shown in Fig.~\ref{fig:analysis}. Similar to \textsl{Vid2Burn\textsubscript{Diverse}} dataset, we use $39$ categories coming from NTU RGBD~\cite{shahroudy2016ntu} dataset to construct the \textsl{Vid2Burn\textsubscript{ADL}} dataset, where the color dot indicates the sample-wise calorie consumption annotation. Compared with \textsl{Vid2Burn\textsubscript{Diverse}} introduced in Fig.~\ref{fig:analysis1}, \textsl{Vid2Burn\textsubscript{ADL}} shows relatively lower movement intensity. In Fig.~\ref{fig:analysis} we can find that for sample-wise calorie consumption annotation, there is an overlapping for fluctuated calorie consumption ranges among different action types.
Finally, we will give a detail description for the heart rate collection procedure.
During the data collection process, each participant needs to wear the wrist band and monitor the heart rate. For a specific action, participants were asked to repeat the action for two minutes and maintain the same action frequency as the original video (randomly selected for each action category leveraged in our work based on NTU RGBD~\cite{shahroudy2016ntu} dataset) to obtain stable heart rate data. The interval between each action is carefully selected to ensure that the heart rate has returned to the rest state heart rate based on measurement.
\begin{table}[t]
\centering
\scalebox{0.8}{
\begin{tabular}{lcccc} 
\toprule
\textbf{Datasets} & \multicolumn{1}{l}{\textbf{Ours-DVS}} & \multicolumn{1}{l}{\textbf{Ours-ADL}} & \multicolumn{1}{l}{\textbf{Stanford-ECM}} & \multicolumn{1}{l}{\textbf{Sphere}} \\ 
\midrule
Modality & Video & Video & Video & video \\
Setting & Fixed- & Fixed-& Ego- & \multicolumn{1}{l}{Fixed-} \\
\#Actions & 33 & 39 & 24 & 11 \\
\#Clips & 4260 & 5529 & 113 & 20 \\
Unit & Calorie & Calorie & MET & MET \\
Label & s /c & s & c & c \\
\bottomrule
\end{tabular}}
\caption{A comparison among datasets for human energy expenditure prediction, where c indicates category-wise annotation and s indicates sample-wise annotation. Ours-DVS indicates the \textsl{Vid2Burn\textsubscript{Diverse}} benchmark and Ours-ADL indicates the \textsl{Vid2Burn\textsubscript{ADL}} benchmark}
\label{tab:comparison}
\vskip-3ex
\end{table}
\section{Supplementary for the experiments and analyses}
First, more details about the category-wise performance for calorie consumption estimation on the \textsl{Vid2Burn\textsubscript{Diverse}} benchmark using category-wise annotation for supervision is represented in Table~\ref{tab:table2}. Second, we provide additional comparison between deep learning-based and pure skeleton-based forward computation for calorie consumption prediction. Finally, we provide an additional ablation studies for different $\sigma$ when generating soft label for supervision.

\begin{table}[t]
\centering
\scalebox{0.75}{
\begin{tabular}{llllllll} 
\toprule
\multirow{2}{*}{\textbf{Method}} & \multicolumn{3}{c}{\textbf{Known activity types}} &  & \multicolumn{3}{c}{\textbf{New activity types}} \\
 & \textcolor[rgb]{1,0.369,0.259}{\textbf{MAE}} & \textbf{SPC} & \textbf{NLL} &  & \textcolor[rgb]{1,0.369,0.259}{\textbf{MAE}} & \textbf{SPC} & \textbf{NLL} \\ 
\midrule
Skeleton (Diverse) & {\cellcolor[rgb]{0.753,0.753,0.753}}330.7& - & - &  & {\cellcolor[rgb]{0.753,0.753,0.753}}665.1 & - & - \\
Skeleton (ADL) & {\cellcolor[rgb]{0.753,0.753,0.753}}78.1 & - & - &  & {\cellcolor[rgb]{0.753,0.753,0.753}}95.2 & - & - \\
SF-AVR (Diverse) & {\cellcolor[rgb]{0.753,0.753,0.753}}57.9 & 49.41 & 6.08 &  & {\cellcolor[rgb]{0.753,0.753,0.753}}134.0 & 42.20 & 9.02 \\
SF-AVR (ADL) & {\cellcolor[rgb]{0.753,0.753,0.753}}20.1 & 68.61 & 5.57 &  & {\cellcolor[rgb]{0.753,0.753,0.753}}36.4 & 72.30 & 6.69 \\
\bottomrule
\end{tabular}}
\caption{A comparison between deep learning-based approach (\emph{I3D-AVR}) and skeleton-based forward computation approach (the same with the skeleton-based calorie consumption annotation generation procedure) on \textsl{Vid2Burn\textsubscript{Diverse}} and \textsl{Vid2Burn\textsubscript{ADL}} benchmarks.}
\vskip-3ex
\label{tab:skeleton}
\end{table}

\subsection{Comparison between deep learning-based approaches and skeleton-based computations}
Since one of the annotation source for calorie consumption estimation is skeleton data, the performance of directly using skeleton to compute calorie consumption is interesting to be researched. We thereby conduct experiments on the two proposed benchmarks with sample-wise annotations between deep learning-based approaches and pure skeleton-based forward calculation. According to the experimental results introduced by Table~\ref{tab:skeleton}, pure skeleton based forward calculation shows a performance difference by $272.8$ kcal and $531.1$ kcal on the known and unknown action types on the \textsl{Vid2Burn\textsubscript{Diverse}} dataset and the performance difference on the \textsl{Vid2Burn\textsubscript{ADL}} dataset for the known and unknown action types are $58.1$ kcal and $58.8$ kcal compared with \emph{I3D-AVR} illustrating the outstanding performance of the deep learning-based approaches for video-based calorie consumption estimation.  
\vspace{-0.2cm}
\subsection{Ablation studies on different standard error for soft label generation}
In order to investigate the influence brought by different $\sigma$ when generating soft label for calorie consumption prediction, we conduct corresponding ablation studies shown in Table~\ref{tab:table8} using \emph{I3D-AVR} approach on the \textsl{Vid2Burn\textsubscript{Diverse}} dataset under category-wise supervision and choose $\sigma$ as 5, 15, 25 and 50 kcal separately. According to the experimental results, choosing $\sigma$ as 15 shows the best performance on known activity types and 50 shows the best performance on unknown activity types in the MAE metric.
\begin{table}[t]
\centering
\scalebox{0.75}{
\begin{tabular}{llllllll} 
\toprule
\multirow{2}{*}{\textbf{Method}} & \multicolumn{3}{c}{\textbf{Known activity types}} &  & \multicolumn{3}{c}{\textbf{New activity types}} \\
 & \textcolor[rgb]{1,0.369,0.259}{\textbf{MAE}} & \textbf{SPC} & \textbf{NLL} &  & \textcolor[rgb]{1,0.369,0.259}{\textbf{MAE}} & \textbf{SPC} & \textbf{NLL} \\ 
\midrule
Video & {\cellcolor[rgb]{0.753,0.753,0.753}}43.2 & 72.82 & 6.19 &  & {\cellcolor[rgb]{0.753,0.753,0.753}}56.0 & 71.08 & \textbf{6.49} \\
I3D-AVR (TFS) & {\cellcolor[rgb]{0.753,0.753,0.753}}102.8 & 43.08 & 6.67 &  & {\cellcolor[rgb]{0.753,0.753,0.753}}69.5 & 52.69 & 6.65 \\
I3D-AVR & {\cellcolor[rgb]{0.753,0.753,0.753}}22.9 & 72.97 & 5.66 &  & {\cellcolor[rgb]{0.753,0.753,0.753}}39.6 & 70.45 & 6.38 \\
\bottomrule
\end{tabular}}
\caption{Experiments on \textsl{Vid2Burn\textsubscript{ADL}} with sample-wise label, where \emph{Video} denotes the model \emph{I3D-AVR} only fine-tuned on the calorie consumption estimation head consisted mainly of fc layers, \emph{I3D-AVR (TFS)} denotes the \emph{I3D-AVR} model while training from scratch.}
\vskip-3ex
\label{tab:table11}
\end{table}
\begin{table}[t]
\centering

\scalebox{0.8}{
\begin{tabular}{llllllll} 
\toprule
\multirow{2}{*}{\textbf{STD}} & \multicolumn{3}{c}{\textbf{Known activity types}} &  & \multicolumn{3}{c}{\textbf{New activity types}} \\
 & \textcolor[rgb]{1,0.369,0.259}{\textbf{MAE}} & \textbf{SPC} & \textbf{NLL} &  & \textcolor[rgb]{1,0.369,0.259}{\textbf{MAE}} & \textbf{SPC} & \textbf{NLL} \\ 

\midrule
50 & {\cellcolor[rgb]{0.753,0.753,0.753}}35.9 & 85.99 & 6.64 &  & {\cellcolor[rgb]{0.753,0.753,0.753}}183.2 & 56.97 & 8.25 \\
25 & {\cellcolor[rgb]{0.753,0.753,0.753}}32.5 & 71.60 & 6.07 &  & {\cellcolor[rgb]{0.753,0.753,0.753}}229.7 & 40.44 & 9.56 \\
15 & {\cellcolor[rgb]{0.753,0.753,0.753}}29.3 & 60.74 & 5.63 &  & {\cellcolor[rgb]{0.753,0.753,0.753}}194.5 & 34.16 & 10.61 \\
5 &  {\cellcolor[rgb]{0.753,0.753,0.753}}38.9 & 32.25 & 4.70 &  & {\cellcolor[rgb]{0.753,0.753,0.753}}228.7 & 7.61 & 16.32 \\
\bottomrule
\end{tabular}}
\caption{Ablation studies by adjusting the $\sigma$ for label softening on \textsl{Vid2Burn\textsubscript{Diverse}} using category-wise label supervision.}
\vskip-3ex
\label{tab:table8}
\end{table}
\begin{table*}
\centering

\scalebox{0.70}{
 \begin{tabular}{llllllllllllllllll} 
\toprule
\multicolumn{1}{c}{\multirow{3}{*}{\textbf{ Method }}} & \multicolumn{10}{c}{\textbf{Known action types (five common classes) }} &  & \multicolumn{6}{c}{\textbf{Unknown action types (Three common classes) }} \\ 
\cline{2-11}\cline{13-18}
\multicolumn{1}{c}{} & \multicolumn{2}{c}{\textbf{Sitting }} & \multicolumn{2}{c}{\textbf{Running }} & \multicolumn{2}{c}{\textbf{Climbing }} & \multicolumn{2}{c}{\textbf{KickBall }} & \multicolumn{2}{c}{\textbf{Punching }} &  & \multicolumn{2}{c}{\textbf{Yoga }} & \multicolumn{2}{c}{\textbf{Sleeping }} & \multicolumn{2}{c}{\textbf{Shopping }} \\
\multicolumn{1}{c}{} & \multicolumn{1}{c}{\textbf{SPC }} & \multicolumn{1}{c}{\textcolor[rgb]{1,0.357,0.161}{\textbf{MAE }}} & \multicolumn{1}{c}{\textbf{SPC }} & \multicolumn{1}{c}{\textcolor[rgb]{1,0.357,0.161}{\textbf{MAE }}} & \multicolumn{1}{c}{\textbf{SPC }} & \multicolumn{1}{c}{\textbf{\textcolor[rgb]{1,0.357,0.161}{MAE }}} & \multicolumn{1}{c}{\textbf{SPC }} & \multicolumn{1}{c}{\textbf{\textcolor[rgb]{1,0.357,0.161}{MAE }}} & \multicolumn{1}{c}{\textbf{SPC }} & \multicolumn{1}{c}{\textbf{\textcolor[rgb]{1,0.357,0.161}{MAE }}} &  & \multicolumn{1}{c}{\textbf{SPC }} & \multicolumn{1}{c}{\textbf{\textcolor[rgb]{1,0.357,0.161}{MAE }}} & \multicolumn{1}{c}{\textbf{SPC }} & \multicolumn{1}{c}{\textbf{\textcolor[rgb]{1,0.357,0.161}{MAE }}} & \multicolumn{1}{c}{\textbf{SPC }} & \multicolumn{1}{c}{\textbf{\textcolor[rgb]{1,0.357,0.161}{MAE }}} \\ 
\hline
ST-GCN & 9.41 & {\cellcolor[rgb]{0.753,0.753,0.753}}245.6 & -13.51 & {\cellcolor[rgb]{0.753,0.753,0.753}}310.9 & 14.81 & {\cellcolor[rgb]{0.753,0.753,0.753}}225.6 & -9.71 & {\cellcolor[rgb]{0.753,0.753,0.753}}554.2 & 10.93 & {\cellcolor[rgb]{0.753,0.753,0.753}}367.5 &  & 19.65 & {\cellcolor[rgb]{0.753,0.753,0.753}}230.5 & -14.38 & {\cellcolor[rgb]{0.753,0.753,0.753}}370.2 & -0.24 & {\cellcolor[rgb]{0.753,0.753,0.753}}322.3 \\ 
\hline
I3D-AVR & 11.09 & {\cellcolor[rgb]{0.753,0.753,0.753}}176.8 & 7.69 & {\cellcolor[rgb]{0.753,0.753,0.753}}82.6 & 13.96 & {\cellcolor[rgb]{0.753,0.753,0.753}}81.8 & 16.51 & {\cellcolor[rgb]{0.753,0.753,0.753}}216.3 & 46.79 & {\cellcolor[rgb]{0.753,0.753,0.753}}1.1 &  & 13.10 & {\cellcolor[rgb]{0.753,0.753,0.753}}191.9 & 1.05 & {\cellcolor[rgb]{0.753,0.753,0.753}}273.2 & 1.92 & {\cellcolor[rgb]{0.753,0.753,0.753}}190.5 \\
SF-AVR & 22.02 & {\cellcolor[rgb]{0.753,0.753,0.753}}88.6 & 11.76 & {\cellcolor[rgb]{0.753,0.753,0.753}}43.8 & 24.19 & {\cellcolor[rgb]{0.753,0.753,0.753}}57.7 & 15.22 & {\cellcolor[rgb]{0.753,0.753,0.753}}202.0 & 30.08 & {\cellcolor[rgb]{0.753,0.753,0.753}}0.9 &  & 20.18 & {\cellcolor[rgb]{0.753,0.753,0.753}}101.9 & 3.37 & {\cellcolor[rgb]{0.753,0.753,0.753}}205.6 & 3.43 & {\cellcolor[rgb]{0.753,0.753,0.753}}174.4 \\
R(2+1)D-AVR & 21.18 & {\cellcolor[rgb]{0.753,0.753,0.753}}124.6 & 12.88 & {\cellcolor[rgb]{0.753,0.753,0.753}}288.0 & 25.03 & {\cellcolor[rgb]{0.753,0.753,0.753}}74.0 & -0.02 & {\cellcolor[rgb]{0.753,0.753,0.753}}31.8 & 32.70 & {\cellcolor[rgb]{0.753,0.753,0.753}}6.3 &  & 15.20 & {\cellcolor[rgb]{0.753,0.753,0.753}}217.9 & 10.97 & {\cellcolor[rgb]{0.753,0.753,0.753}}314.2 & 4.68 & {\cellcolor[rgb]{0.753,0.753,0.753}}271.1 \\
R3D-AVR & 18.14 & {\cellcolor[rgb]{0.753,0.753,0.753}}166.5 & 6.14 & {\cellcolor[rgb]{0.753,0.753,0.753}}102.3 & 21.13 & {\cellcolor[rgb]{0.753,0.753,0.753}}135.5 & 2.87 & {\cellcolor[rgb]{0.753,0.753,0.753}}246.4 & 38.84 & {\cellcolor[rgb]{0.753,0.753,0.753}}0.2 &  & 17.82 & {\cellcolor[rgb]{0.753,0.753,0.753}}97.4 & 2.71 & {\cellcolor[rgb]{0.753,0.753,0.753}}168.3 & 1.17 & {\cellcolor[rgb]{0.753,0.753,0.753}}108.5 \\
I3D-LSTM & 14.60 & {\cellcolor[rgb]{0.753,0.753,0.753}}74.7 & 6.09 & {\cellcolor[rgb]{0.753,0.753,0.753}}147.7 & 14.85 & {\cellcolor[rgb]{0.753,0.753,0.753}}99.9 & 10.15 & {\cellcolor[rgb]{0.753,0.753,0.753}}250.0 & 48.97 & {\cellcolor[rgb]{0.753,0.753,0.753}}0.6 &  & 12.04 & {\cellcolor[rgb]{0.753,0.753,0.753}}217.6 & 0.14 & {\cellcolor[rgb]{0.753,0.753,0.753}}287.2 & 4.15 & {\cellcolor[rgb]{0.753,0.753,0.753}}280.0 \\
SF-LSTM & 15.02 & {\cellcolor[rgb]{0.753,0.753,0.753}}182.0 & 12.59 & {\cellcolor[rgb]{0.753,0.753,0.753}}9.6 & 15.79 & {\cellcolor[rgb]{0.753,0.753,0.753}}86.7 & 8.69 & {\cellcolor[rgb]{0.753,0.753,0.753}}161.4 & 45.81 & {\cellcolor[rgb]{0.753,0.753,0.753}}7.6 &  & 19.49 & {\cellcolor[rgb]{0.753,0.753,0.753}}133.6 & 0.26 & {\cellcolor[rgb]{0.753,0.753,0.753}}373.7 & 3.91 & {\cellcolor[rgb]{0.753,0.753,0.753}}266.8 \\
R(2+1)D-LSTM & 8.97 & {\cellcolor[rgb]{0.753,0.753,0.753}}218.5 & 11.22 & {\cellcolor[rgb]{0.753,0.753,0.753}}37.1 & 10.47 & {\cellcolor[rgb]{0.753,0.753,0.753}}98.4 & 9.86 & {\cellcolor[rgb]{0.753,0.753,0.753}}347.1 & 45.54 & {\cellcolor[rgb]{0.753,0.753,0.753}}5.3 &  & 18.26 & {\cellcolor[rgb]{0.753,0.753,0.753}}181.8 & 5.71 & {\cellcolor[rgb]{0.753,0.753,0.753}}152.6 & 8.53 & {\cellcolor[rgb]{0.753,0.753,0.753}}134.4 \\
R3D-LSTM & 8.78 & {\cellcolor[rgb]{0.753,0.753,0.753}}128.3 & 6.54 & {\cellcolor[rgb]{0.753,0.753,0.753}}262.1 & 8.82 & {\cellcolor[rgb]{0.753,0.753,0.753}}83.8 & 6.48 & {\cellcolor[rgb]{0.753,0.753,0.753}}204.5 & 41.82 & {\cellcolor[rgb]{0.753,0.753,0.753}}3.4 &  & 15.66 & {\cellcolor[rgb]{0.753,0.753,0.753}}223.2 & -0.04 & {\cellcolor[rgb]{0.753,0.753,0.753}}340.9 & 0.02 & {\cellcolor[rgb]{0.753,0.753,0.753}}370.1 \\
\bottomrule
\end{tabular}}
\caption{Experimental results for human calorie consumption estimation for the selected action categories on the \textsl{Vid2Burn\textsubscript{Diverse}} dataset supervised with category-wise annotation.}
\label{tab:table2}
\end{table*}
\subsection{Further illustration of the calorie consumption estimation ability} When digging deeper into the direction of the deep learning-based calorie consumption estimation, the relationship between action recognition and calorie consumption estimation is interesting to be investigated, especially for the question about whether there is only a lookup relationship between calorie consumption estimation and action recognition or not. First if looking into labels, we have sample-wise label differing among the samples inside same action type according to different human body movement intensity, which makes sure that it will not be a simple lookup relationship.
According to Fig.~\ref{fig:analysis}, there are calorie consumption range overlapping among different action types.
Second we conduct several ablation studies listed in Table \ref{tab:table11} to support our argument. If our models predict lookup relationship between calorie consumption and action classes, the performance of the model only fine-tuning the fc layers should be higher than the performance of our approach.
Since video classes are highly dependent on action classes, we conduct experiment by freezing weights of pretrained video-based backbone while only adjusting weights of fully-connected layers as \emph{Video} in Table \ref{tab:table11}, where MAE of \emph{Video} for both known- and unknown-action types evaluation are all worse than \emph{I3D-AVR}. We also test train-from-scratch for the \emph{I3D-AVR} baseline denoted as \emph{I3D-AVR (TFS)} which shows the worst performance when compared with others, illustrating that pretraining is important.
Through the above analyses it can be seen that the relationship between human action and calorie consumption prediction is not a simple lookup relationship and also pretraining is essential.
\subsection{Supplementary for implementation details}
In addition to the mentioned implementation details in our paper, our model is built based on PyTorch toolbox. 
Since we leverage temporal sliding window to aggregate features along time axis, the temporal overlapping of the sliding window for I3D, R3D, R(2+1)D backbones are chosen as 6 frames while the temporal overlapping for SlowFast is chosen as 16 frames since it requires larger temporal window length (32 frames) compared with the others (16 frames).
For the \textsl{Vid2Burn\textsubscript{ADL}} dataset, the estimation head has 500 channels output as the maximum calorie consumption estimation range is set as 500 kcal together with resolution 1 kcal.
For the \textsl{Vid2Burn\textsubscript{Diverse}} dataset, the channel number of the final output is 1000.

\end{document}